\def\year{2022}\relax
\documentclass[letterpaper]{article} 
\usepackage{aaai22}  
\usepackage{times}  
\usepackage{helvet}  
\usepackage{courier}  
\usepackage[hyphens]{url}  
\usepackage{graphicx} 
\usepackage{natbib}  
\usepackage{caption} 
\DeclareCaptionStyle{ruled}{labelfont=normalfont,labelsep=colon,strut=off} 
\usepackage{algorithm}
\usepackage{algorithmic}

\usepackage{newfloat}
\usepackage{listings}
\floatstyle{ruled}
\newfloat{listing}{tb}{lst}{}
\floatname{listing}{Listing}

\usepackage[switch]{lineno}
\usepackage{amsmath}
\usepackage{amssymb}
\usepackage{comment}
\usepackage{subcaption}
\DeclareMathAlphabet\mathbfcal{OMS}{cmsy}{b}{n}
\newcommand{\round}[1]{\ensuremath{\lfloor#1\rceil}}

\usepackage{adjustbox}
\usepackage{nicefrac}
\usepackage{booktabs}
\usepackage{comment}
\usepackage{multirow}
\usepackage[dvipsnames]{xcolor}
\usepackage{pdfpages}

\title{Towards Light-weight and Real-time Line Segment Detection}
\author {

        Geonmo Gu\thanks{Authors contributed equally.},
        Byungsoo Ko\footnotemark[1],
        SeoungHyun Go,
        Sung-Hyun Lee,
        Jingeun Lee,
        Minchul Shin\\
}
\affiliations {
    NAVER/LINE Vision\\
    \{korgm403, kobiso62, powflash, shlee.mars, jglee0206, min.stellastra\}@gmail.com
}

\begin{document}

\maketitle

\begin{abstract}
Previous deep learning-based line segment detection (LSD) suffers from the immense model size and high computational cost for line prediction.
This constrains them from real-time inference on computationally restricted environments.
In this paper, we propose a real-time and light-weight line segment detector for resource-constrained environments named Mobile LSD (M-LSD).
We design an extremely efficient LSD architecture by minimizing the backbone network and removing the typical multi-module process for line prediction found in previous methods.
To maintain competitive performance with a light-weight network, we present novel training schemes: Segments of Line segment (SoL) augmentation, matching and geometric loss.
SoL augmentation splits a line segment into multiple subparts, which are used to provide auxiliary line data during the training process.
Moreover, the matching and geometric loss allow a model to capture additional geometric cues.
Compared with TP-LSD-Lite, previously the best real-time LSD method, our model (M-LSD-tiny) achieves competitive performance with 2.5\% of model size and an increase of 130.5\% in inference speed on GPU.
Furthermore, our model runs at 56.8 FPS and 48.6 FPS on the latest Android and iPhone mobile devices, respectively.
To the best of our knowledge, this is the first real-time deep LSD available on mobile devices.
Our code is available~\footnote{https://github.com/navervision/mlsd}.
\end{abstract}

\section{Introduction}

Line segments and junctions are crucial visual features in low-level vision, which provide fundamental information to the higher level vision tasks, such as pose estimation~\cite{pvribyl2017absolute,xu2016pose}, structure from motion~\cite{bartoli2005structure,micusik2017structure}, 3D reconstruction~\cite{denis2008efficient, faugeras1992depth}, image matching~\cite{xue2017anisotropic}, wireframe to image translation~\cite{xue2019neural} and image rectification~\cite{xue2019rectification}.
Moreover, the growing demand for performing such vision tasks on resource constraint platforms, like mobile or embedded devices, has made real-time line segment detection (LSD) an essential but challenging task. 
The difficulty arises from the limited computational power and model size when finding the best accuracy and resource-efficiency trade-offs to achieve real-time inference.

\begin{figure}[t!]
\centering
\includegraphics[width=0.8\columnwidth]{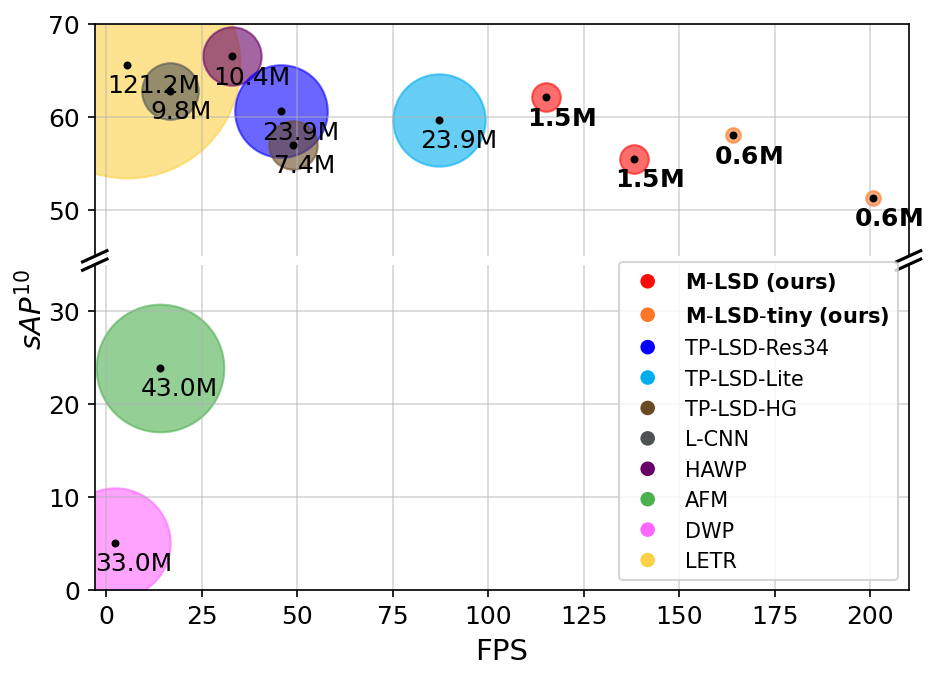}
\caption{Comparison of M-LSD and existing LSD methods on Wireframe dataset. Inference speed (FPS) is computed on Tesla V100 GPU. Size and value of circles indicate the number of model parameters (Millions). M-LSD achieves competitive performance with the lightest model size and the fastest inference speed. Details are in Table~\ref{table:sota}.}
\label{fig:teaser}
\end{figure}

\begin{figure*}[t!]
\includegraphics[width=0.9\textwidth]{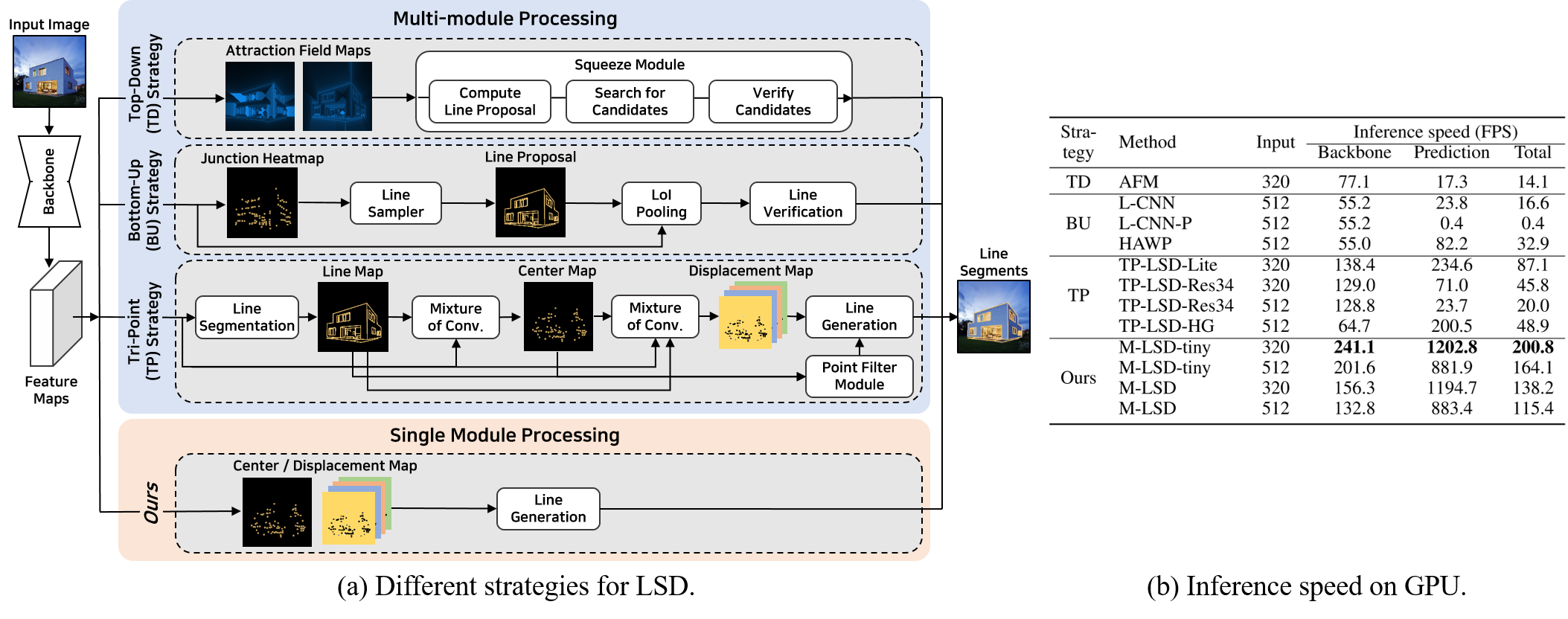}
\caption{(a) Previous LSD methods exploit multi-module processing for line segment prediction. In contrast, our method directly predicts line segments from feature maps with a single module. (b) Our method shows superior speed on backbone and line prediction by employing a light-weight network with a single module of line prediction.}%
\label{fig:strategy}
\end{figure*}

With the advent of deep neural networks, deep learning-based LSD architectures have adopted models to learn various geometric cues of line segments and have proved to show improvements in performance.
As described in Figure~\ref{fig:strategy}, we have summarized multiple strategies that use deep learning models for LSD.
The top-down strategy~\cite{xue2019learning} first detects regions of line segment with attraction field maps and then squeezes these regions into line segments to make predictions.
In contrast, the bottom-up strategy first detects junctions, then arranges them into line segments, and lastly verifies the line segments by using an extra classifier~\cite{zhou2019end,xue2020holistically,zhang2019ppgnet} or a merging algorithm~\cite{huang2019wireframe,huang2018learning}.
Recently, \cite{huang2020tp} proposes Tri-Points (TP) representation for a simpler process of line prediction without the time-consuming steps of line proposal and verification.

Although previous efforts of using deep networks have made remarkable achievements, real-time inference for LSD on resource-constraint platforms still remains limited.
There have been attempts to present real-time LSD~\cite{huang2020tp,meng2020lgnn,xue2020holistically}, but they still depend on server-class GPUs.
This is mainly because the models that are used exploit heavy backbone networks, such as dilated ResNet50-based FPN~\cite{zhang2019ppgnet}, stacked hourglass network~\cite{meng2020lgnn,huang2020tp}, and atrous residual U-net~\cite{xue2019learning}, which require large memory and high computational power.
In addition, as shown in Figure~\ref{fig:strategy}, the line prediction process consists of multiple modules, which include line proposal~\cite{xue2019learning,zhang2019ppgnet,zhou2019end,xue2020holistically}, line verification networks~\cite{zhang2019ppgnet,zhou2019end,xue2020holistically} and mixture of convolution module~\cite{huang2020tp,huang2018learning}.
As the size of the model and the number of modules for line prediction increase, the overall inference speed of LSD can become slower, as shown in Figure~\ref{fig:strategy}b, while demanding higher computation.
Thus, increases in computational cost make it difficult to deploy LSD on resource-constraint platforms.

In this paper, we propose a real-time and light-weight LSD for resource-constrained environments, named Mobile LSD (M-LSD).
For the network, we design a significantly efficient architecture with a single module to predict line segments.
By minimizing the network size and removing the multi-module process from previous methods, M-LSD is extremely light and fast.
To maintain competitive performance even with a light-weight network, we present novel training schemes: SoL augmentation, matching and geometric loss.
SoL augmentation divides a line segment into subparts, which are further used to provide augmented line data during the training phase.
Matching and geometric loss train a model with additional geometric information, including relation between line segments, junction and line segmentation, length and degree regression.
As a result, our model is able to capture extra geometric information during training to make more accurate line predictions.
Moreover, the proposed training schemes can be used with existing methods to further improve performance in a plug-and-play manner.

As shown in Figure~\ref{fig:teaser}, our methods achieve competitive performance and faster inference speed with a much smaller model size.
M-LSD outperforms previously the real-time method, TP-LSD-Lite~\cite{huang2020tp}, with only 6.3\% of the model size but gaining an increase of 32.5\% in inference speed.
Moreover, M-LSD-tiny runs in real-time at 56.8 FPS and 48.6 FPS on the latest Android and iPhone mobile devices, respectively.
To the best of our knowledge, this is the first real-time LSD method available on mobile devices.

\section{Related Works}

\begin{figure*}[t!]
\centering
\includegraphics[width=0.78\textwidth]{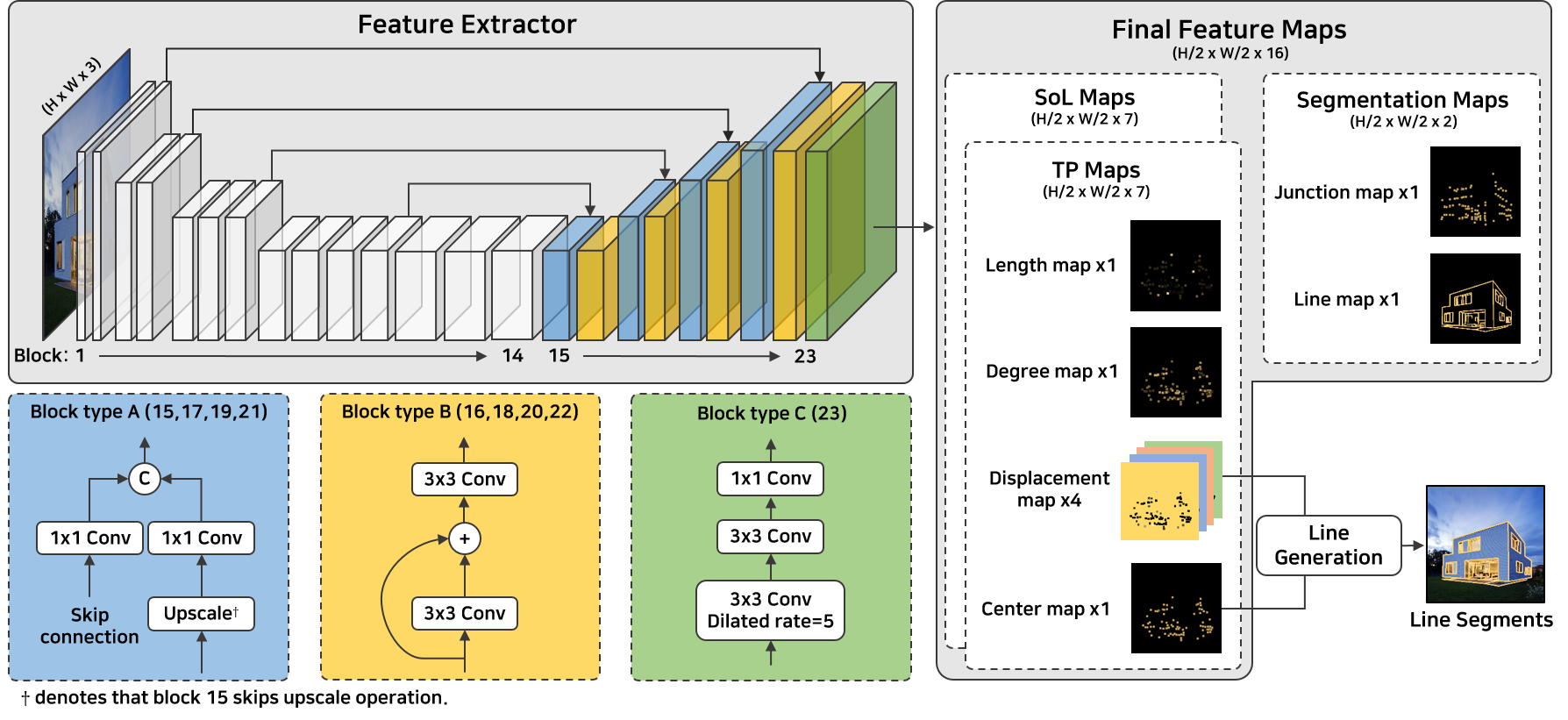}
\caption{The overall architecture of M-LSD. In the feature extractor, block 1 $\sim$ 14 are parts of MobileNetV2, and block 15 $\sim$ 23 are designed as a top-down architecture. The predicted line segments are generated with center and displacement maps.}
\label{fig:framework}
\end{figure*}

\textbf{Deep Line Segment Detection. }
There have been active studies on deep learning-based LSD.
In junction-based methods, DWP~\cite{huang2018learning} includes two parallel branches to predict line and junction heatmaps, followed by a merging process.
PPGNet~\cite{zhang2019ppgnet} and L-CNN~\cite{zhou2019end} utilize junction-based line segment representations with an extra classifier to verify whether a pair of points belongs to the same line segment.
Another approach uses dense prediction.
AFM~\cite{xue2019learning} predicts attraction field maps that contain 2-D projection vectors representing associated line segments, followed by a squeeze module to recover line segments.
HAWP~\cite{xue2020holistically} is presented as a hybrid model of AFM and L-CNN.
Recently, \cite{huang2020tp} devises the TP line representation to remove the use of extra classifiers or heuristic post-processing found in previous methods and proposes TP-LSD network with two branches: TP extraction and line segmentation branches.
Other approaches include the use of transformers~\cite{xu2021line} or Hough transform with deep networks~\cite{lin2020deep}.
However, it is commonly observed that the aforementioned multi-module processes restrict existing LSD to run on resource-constrained environments.

\textbf{Real-time Object Detectors. }
Real-time object detection has been an important task for deep learning-based object detection.
Object detectors proposed in earlier days, such as RCNN-series~\cite{girshick2014rcnn,girshick2015fast,ren2015faster}, consist of two-stage architecture: generating proposals in the first stage, then classifying the proposals in the second stage.
These two-stage detectors typically suffer from slow inference speed and difficulty in optimization.
To handle this problem, one-stage detectors, such as YOLO-series~\cite{redmon2016you,redmon2017yolo9000,redmon2018yolov3} and SSD~\cite{liu2016ssd}, are proposed to achieve GPU real-time inference by reducing backbone size and simplifying the two-stage process into one.
This one-stage architecture has been further studied and improved to run in real-time on mobile devices~\cite{howard2017mobilenets,sandler2018mobilenetv2,wang2018pelee,li2018tiny}.
Motivated by the transition from two-stage to one-stage architecture in object detection, we argue that the complicated multi-module processing in previous LSD can be disregarded.
We simplify the line prediction process with a single module for faster inference speed and enhance the performance by the efficient training strategies; SoL augmentation, matching and geometric loss.


\section{M-LSD for Line Segment Detection}
In this section, we present the details of M-LSD.
Our design mainly focuses on efficiency while retaining competitive performance.
Firstly, we exploit a light-weight backbone and reduce the modules involved in processing line predictions for better efficiency.
Next, we apply additional training schemes, including SoL augmentation, matching and geometric loss, to capture extra geometric cues.
As a result, M-LSD is able to balance the trade-off between accuracy and efficiency to be well suited for mobile devices.

\subsection{Network Architecture}

We design light (M-LSD) and lighter (M-LSD-tiny) models as popular encoder-decoder architectures.
In efforts to build a light-weight LSD model, our encoder networks are based on MobileNetV2~\cite{sandler2018mobilenetv2} which is well-known to run in real-time on mobile environments.
The encoder network uses parts of MobileNetV2 to make it even lighter.
As illustrated in Figure~\ref{fig:framework}, the encoder of M-LSD includes an input to 96-channel of bottleneck blocks.
The number of parameters in the encoder network is 0.56M (16.5\% of MobileNetV2), while the total parameters of MobileNetV2 are 3.4M.
For M-LSD-tiny, a slightly smaller yet faster model, the encoder network also uses parts of MobileNetV2, including an input to 64-channel of bottleneck blocks which results in a number of 0.25M (7.4\% of MobileNetV2).
The decoder network is designed using a combination of block types A, B, and C.
The expansive path consists of concatenation of feature maps from the skip connection and upscale from block type A, followed by two 3 $\times$ 3 convolutions with a residual connection in-between from block type B.
Similarly, block type C performs two 3 $\times$ 3 convolutions, the first being a dilated convolution, followed by a 1 $\times$ 1 convolution.
Please refer to the supplementary material for further details on the network architectures.

As shown in Figure~\ref{fig:strategy}b, we observe that one of the most critical bottlenecks in inference speed has been the prediction process, which contains multi-module processing from previous methods.
In this paper, we argue that the complicated multi-module can be disregarded.
As illustrated in Figure~\ref{fig:framework}, we generate line segments directly from the final feature maps in a single module process.
In the final feature maps, each feature map channel serves its own purpose:
1) TP maps have seven feature maps, including one length map, one degree map, one center map, and four displacement maps.
2) SoL maps have seven feature maps with the same configuration as TP maps.
3) Segmentation maps have two feature maps, including junction and line maps.

\begin{figure}[t!]
    \centering
     \begin{subfigure}[b]{0.48\columnwidth}
        \centering\includegraphics[width=0.8\columnwidth]{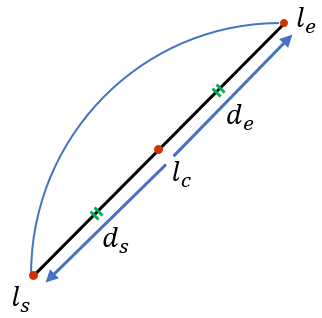}
         \caption{TP representation}
         \label{fig:tp}
     \end{subfigure}
    \begin{subfigure}[b]{0.48\columnwidth}
        \centering\includegraphics[width=0.8\columnwidth]{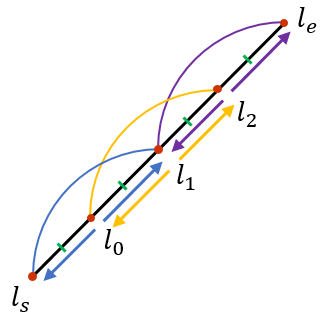}
         \caption{SoL augmentation}
         \label{fig:sol}
     \end{subfigure}
\caption{Tri-Points (TP) representation and Segments of Line segment (SoL) augmentation. $l_s$, $l_c$, and $l_e$ denote start, center, and end points, respectively. $d_s$ and $d_e$ are displacement vectors to start and end points. $l_0 \sim l_2$ indicates internally dividing points of the line segment $\overline{l_s l_e}$.}
\label{fig:line_representation}
\end{figure}

\subsection{Line Segment Representation}
Line segment representation determines how line segment predictions are generated and ultimately affects the efficiency of LSD.
Hence, we employ the TP representation~\cite{huang2020tp} which has been introduced to have a simple line generation process and shown to perform real-time LSD using GPUs.
TP representation uses three key-points to depict a line segment: start, center, and end points.
As illustrated in Figure~\ref{fig:tp}, the start $l_s$ and end $l_e$ points are represented by using two displacement vectors ($d_s$, $d_e$) with respect to the center $l_c$ point.
The line generation process, which is to convert center point and displacement vectors to a vectorized line segment, is performed as:
\begin{eqnarray}
(x_{l_s}, y_{l_s}) = (x_{l_c}, y_{l_c}) + d_s(x_{l_c}, y_{l_c}), \nonumber \\
(x_{l_e}, y_{l_e}) = (x_{l_c}, y_{l_c}) + d_e(x_{l_c}, y_{l_c}),
\label{eq:line_gen}
\end{eqnarray}
where ($x_\alpha$, $y_\alpha$) denotes coordinates of an arbitrary $\alpha$ point.
$d_s(x_{l_c}, y_{l_c})$ and $d_e(x_{l_c}, y_{l_c})$ indicate 2D displacements from the center point $l_c$ to the corresponding start $l_s$ and end $l_e$ points.
The center point and displacement vectors are trained with one center map and four displacement maps (one for each $x$ and $y$ value of the displacement vectors $d_s$ and $d_e$).
In the line generation process, we extract the exact center point position by applying non-maximum suppression on the center map.
Next, we generate line segments with the extracted center points and the corresponding displacement vectors using a simple arithmetic operation as expressed in Equation~\ref{eq:line_gen}; thus, making inference efficient and fast.

\subsection{Matching Loss}
Following~\cite{huang2020tp}, we use the weighted binary cross-entropy (WBCE) loss and smooth L1 loss as center loss $\mathcal L_{center}$ and displacement loss $\mathcal L_{disp}$, which are for training the center and displacement map, respectively.
The line segments under the TP representation are decoupled into center points and displacement vectors, which are optimized separately.
However, the coupled information of the line segment is under-utilized in the objective functions.

\begin{figure}[t!]
    \centering
    \begin{subfigure}[b]{0.48\columnwidth}
        \centering\includegraphics[width=1.0\columnwidth]{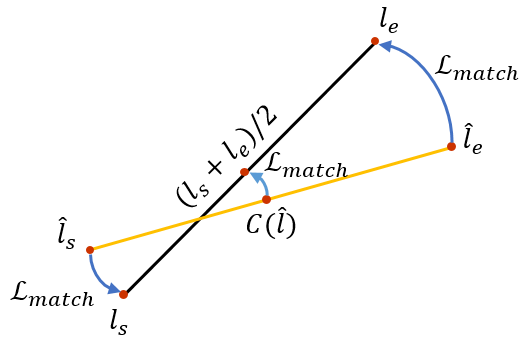}
         \caption{Matching loss}
         \label{fig:matching}
     \end{subfigure}
     \begin{subfigure}[b]{0.48\columnwidth}
        \centering\includegraphics[width=0.8\columnwidth]{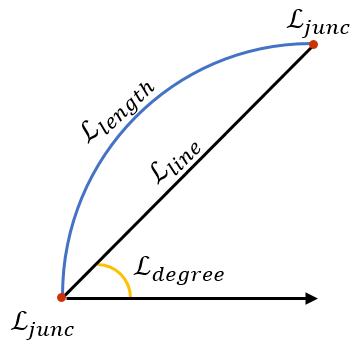}
         \caption{Geometric loss}
         \label{fig:geometric}
     \end{subfigure}
\caption{Matching and geometric loss. (a) Given a matched pair of a predicted line $\hat l$ and a GT line $l$, matching loss ($\mathcal L_{match}$) optimizes the predicted start, end, and center points. (b) Given a line segment, M-LSD learns various geometric cues: junction ($\mathcal L_{junc}$) and line ($\mathcal L_{line}$) segmentation, length ($\mathcal L_{length}$) and degree ($\mathcal L_{degree}$) regression.}
\label{fig:losses}
\end{figure}

To resolve this problem, we present a matching loss, which leverages the coupled information w.r.t. the ground truth.
As illustrated in Figure~\ref{fig:matching}, matching loss considers relation between line segments by guiding the generated line segments to be similar to the matched GT.
We first take the endpoints of each prediction, which can be calculated via the line generation process, and measure the Euclidean distance $d(\cdot)$ to the endpoints of the GT.
Next, these distances are used to match predicted line segments $\hat l$ with GT line segments $l$ that are under a threshold $\gamma$:
\begin{eqnarray}
d(l_s, \hat{l}_s) < \gamma \text{ and } d(l_e, \hat{l}_e) < \gamma,
\end{eqnarray}
where $l_s$ and $l_e$ are the start and end points of the line $l$, and $\gamma$ is set to 5 pixels.
Then, we obtain a set $\mathbb{M}$ of matched line segments ($l$, $\hat l$) that satisfies this condition.
Finally, the L1 loss is used for the matching loss, which aims to minimize the geometric distance of the matched line segments w.r.t the start, end, and center points as follows:
\begin{eqnarray}
\mathcal L_{match}&=&\frac{1}{\mid\mathbb{M}\mid}\sum_{(l,\hat{l}) \in \mathbb{M}} \parallel l_s-\hat{l}_s \parallel_1 + \parallel l_e-\hat{l}_e \parallel_1 \nonumber \\
&&+\parallel \tilde C(\hat{l})-(l_s + l_e)/2 \parallel_1,
\end{eqnarray}
where $\tilde C(\hat{l})$ is the center point of line $\hat{l}$ from the center map.
The total loss function for the TP map can be formulated as $\mathcal L_{TP} =  \mathcal L_{center}  + \mathcal L_{disp} + \mathcal L_{match}$.

\subsection{SoL Augmentation}
We propose Segments of Line segment (SoL) augmentation that increases the number of line segments with wider varieties of length for training.
Learning line segments with center points and displacement vectors can be insufficient in certain circumstances where a line segment may be too long to manage within the receptive field size or the center points of two distinct line segments may be too close to each other.
To address these issues and provide auxiliary information to the TP representation, SoL explicitly splits line segments into multiple subparts with overlapping portions of each other.
An overlap between each split is enforced to preserve connectivity among the subparts.

As described in Figure~\ref{fig:sol}, we compute $k$ internally dividing points ($l_0, l_1, \cdots, l_k$) and separate the line segment $\overline{l_sl_e}$ into $k$ subparts ($\overline{l_sl_1}$, $\overline{l_0l_2}$, $\cdots$, $\overline{l_{k-1}l_e}$).
Expressed in TP representation, each subpart is trained as if it is a typical line segment.
The number of internally dividing points $k$ is determined by the length of the line segment as $k = \round{r(l)/(\mu/2)} - 1$, where $r(l)$ denotes the length of line segment $l$, and $\mu$ is the base length of subparts.
Note that when $k \leq 1$, we do not split the line segment.
The resulting length of each subpart can be similar to $\mu$ with small margins of error due to the rounding function \round{\cdot}, and we empirically set $\mu= input\_size \times 0.125$.
The loss function of $\mathcal L_{SoL}$ follows the same configuration as $\mathcal L_{TP}$, while each subpart is treated as an individual line segment.
Note that the line generation process is only done in TP maps, not in SoL maps.

\begin{table}[t!]
\centering
\begin{adjustbox}{width=1.0\columnwidth,center}
\begin{tabular}{clccc}
\toprule
M & Schemes                & $F^H$ & $sAP^{10}$ & $LAP$ \\
\midrule
1 & Baseline    &   74.3   & 48.9  &  48.1  \\
2 & + Matching loss  &   75.4 \textcolor{Green}{(+1.1)}  & 52.2 \textcolor{Green}{(+3.3)}  &  52.5 \textcolor{Green}{(+4.4)} \\
3 & + Geometric loss     &   76.2 \textcolor{Green}{(+0.8)}  & 55.1 \textcolor{Green}{(+2.9)} &   55.3 \textcolor{Green}{(+2.8)}  \\
4 & + SoL augmentation      &   \textbf{77.2} \textcolor{Green}{(+1.0)}  & \textbf{58.0} \textcolor{Green}{(+2.9)} &   \textbf{57.9} \textcolor{Green}{(+2.6)}  \\
\bottomrule
\end{tabular}
\end{adjustbox}
\caption{Ablation study of M-LSD-tiny on Wireframe. The baseline is M-LSD-tiny trained with only TP representation. M denotes model number.}
\label{table:ablation}
\end{table}

\subsection{Learning with Geometric Information}
To boost the quality of predictions, we incorporate various geometric information about line segments which helps the overall learning process.
In this section, we present learning LSD with junction and line segmentation, and length and degree regression for additional geometric information.

\subsubsection{Junction and Line Segmentation}
Center point and displacement vectors are highly related to pixel-wise junctions and line segments in the segmentation maps of Figure~\ref{fig:framework}.
For example, end points, derived from the center point and displacement vectors, should be the junction points.
Also, center points must be localized on the pixel-wise line segment.
Thus, learning the segmentation maps of junctions and line segments works as a spatial attention cue for LSD.
As illustrated in Figure~\ref{fig:framework}, M-LSD contains segmentation maps, including a junction map and a line map.
We construct the junction GT map by scaling with Gaussian kernel as the center map, while using a binary map for line GT map.
The total segmentation loss is defined as $\mathcal L_{seg} =  \mathcal L_{junc}  + \mathcal L_{line}$, where we use WBCE loss for both $\mathcal L_{junc}$ and $\mathcal L_{line}$.

\subsubsection{Length and Degree Regression}
As displacement vectors can be derived from the length and degree of line segments, they can be additional geometric cues to support the displacement maps.
We compute the length and degree from the ground truth and mark the values on the center of line segments in each GT map.
Next, these values are extrapolated to a $3 \times 3$ window so that all neighboring pixels of a given pixel contain the same value.
As shown in Figure~\ref{fig:framework}, we maintain predicted length and degree maps for both TP and SoL maps, where TP uses the original line segment and SoL uses augmented subparts.
As the ranges of length and degree are wide, we divide each length by the diagonal length of the input image for normalization.
For degree, we divide each degree by $2\pi$ and add 0.5.
The total regression loss can be formulated as $\mathcal L_{reg} = \mathcal L_{length} + \mathcal L_{degree}$, where we use smooth L1 loss for both $\mathcal L_{length}$ and $\mathcal L_{degree}$.

\subsection{Final Loss Functions}
The geometric loss function is defined as the sum of segmentation and regression loss:
\begin{eqnarray}
\label{eq:geo}
\mathcal L_{Geo} = \mathcal L_{seg} + \mathcal L_{reg}.
\end{eqnarray}
The loss function for SoL maps $\mathcal L_{SoL}$ follows the same formulation as $\mathcal L_{TP}$ but with SoL augmented GT.
Finally, we obtain the final loss function to train M-LSD as follows:
\begin{eqnarray}
\label{eq:total}
\mathcal L_{total} &=& \mathcal L_{TP} + \mathcal L_{SoL} + \mathcal L_{Geo}.
\end{eqnarray}
Please refer to the supplementary material for further details on the feature maps and losses.

\begin{figure}[t!]
    \centering
     \begin{subfigure}[b]{0.48\columnwidth}
        \centering\includegraphics[width=0.55\columnwidth]{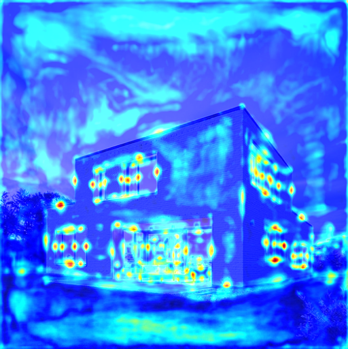}
         \caption{Baseline (M1)}
         \label{fig:match_x}
     \end{subfigure}
    \begin{subfigure}[b]{0.48\columnwidth}
        \centering\includegraphics[width=0.55\columnwidth]{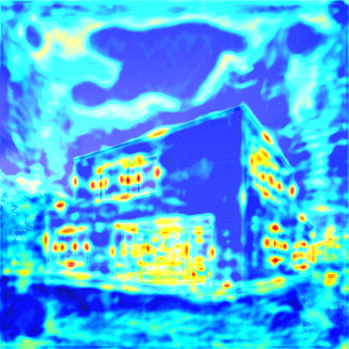}
         \caption{w/ matching loss (M2)}
         \label{fig:match_o}
     \end{subfigure}
     \begin{subfigure}[b]{0.48\columnwidth}
        \centering\includegraphics[width=0.55\columnwidth]{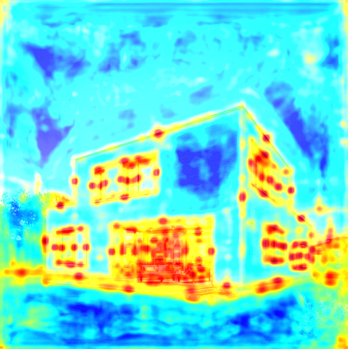}
         \caption{w/ geometric loss (M3)}
         \label{fig:sol_x}
     \end{subfigure}
    \begin{subfigure}[b]{0.48\columnwidth}
        \centering\includegraphics[width=0.55\columnwidth]{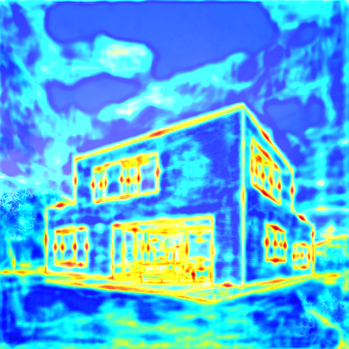}
         \caption{w/ SoL augmentation (M4)}
         \label{fig:sol_o}
     \end{subfigure}
\caption{Saliency maps generated from TP center map. Model numbers (M1$\sim$4) are from Table~\ref{table:ablation}.}
\label{fig:comparison_cam}
\end{figure}

\section{Experiments}
In this section, we conduct extensive ablation studies, quantitative and qualitative analysis of the proposed method.
For better understanding, we add extended experiments in the supplementary material, including ablation study of architecture, SoL augmentation, application example and so on.

\begin{table*}[t!]
\centering
\begin{adjustbox}{width=0.8\textwidth,center}
\begin{tabular}{lcccccccccccc}
\toprule
\multirow{2}{*}{Methods} & \multirow{2}{*}{Input} & \multicolumn{4}{c}{Wireframe} & \multicolumn{4}{c}{YorkUrban} & \multirow{2}{*}{Params(M)} & \multirow{2}{*}{FPS} \\ \cline{3-10}
                         &                        & F$^H$   & sAP$^5$ & sAP$^{10}$ & LAP  & F$^H$   & sAP$^5$ & sAP$^{10}$ & LAP  &                            &                      \\
\midrule
LSD~\cite{von2008lsd}    & 320                    & 64.1  & 6.7   & 8.8    & 18.7 & 60.6  & 7.5   & 9.2    & 16.1 & -                           & 100.0$^{\dagger}$                \\
DWP~\cite{huang2018learning}& 512                    & 72.7  & 3.7     & 5.1    & 6.6  & 65.2  & 2.8     & 2.6    & 3.1  & 33.0                       & 2.2                  \\
AFM~\cite{xue2019learning}& 320                    & 77.3  & 18.3  & 23.9   & 36.7 & 66.3  & 7.0   & 9.1    & 17.5 & 43.0                       & 14.1                 \\
LGNN~\cite{meng2020lgnn} & 512                    & -     & -     & 62.3   & -    & -     & -     & -      & -    & -                           & 15.8$^{\ddagger}$                 \\
LGNN-lite~\cite{meng2020lgnn}& 512                    & -     & -     & 57.6   & -    & -     & -     & -      & -    & -                          & 34.0$^{\ddagger}$                 \\
TP-LSD-Lite~\cite{huang2020tp}& 320                    & 80.4  & 56.4  & 59.7   & 59.7 & \textcolor{blue}{\textbf{68.1}}  & 24.8  & 26.8   & 31.2 & 23.9                       & \textcolor{blue}{87.1}                 \\
TP-LSD-Res34~\cite{huang2020tp}& 320                    & 81.6  & 57.5  & 60.6   & 60.6 & 67.4  & 25.3  & 27.4   & 31.1 & 23.9                       & 45.8                 \\
TP-LSD-Res34~\cite{huang2020tp}& 512                    & 80.6  & 57.6  & 57.2   & 61.3 & 67.2  & \textcolor{blue}{\textbf{27.6}}  & 27.7   & \textcolor{blue}{\textbf{34.3}} & 23.9                       & 20.0                 \\
TP-LSD-HG~\cite{huang2020tp}& 512                    & 82.0  & 50.9  & 57.0   & 55.1 & 67.3  & 18.9  & 22.0   & 24.6 & \textcolor{blue}{7.4}                        & 48.9                 \\
LETR~\cite{xu2021line}& 1100$^{\ast}$                    & \textcolor{blue}{\textbf{82.6}}  & 59.2  & 65.6   & \textcolor{blue}{\textbf{65.1}} & 66.6  & 24.0  & 27.6   & 32.5 & 121.2                        & 5.4                 \\
L-CNN~\cite{zhou2019end} & 512                    & 77.5  & 58.9  & 62.8   & 59.8 & 64.6  & 25.9  & 28.2   & 32.0 & 9.8                        & 16.6                 \\
HAWP~\cite{xue2020holistically}& 512                    & 80.3  & 62.5  & 66.5   & 62.9 & 64.8  & 26.1  & \textcolor{blue}{\textbf{28.5}}   & 30.4 & 10.4                       & 32.9                 \\
HT-L-CNN~\cite{lin2020deep}& 512                    & -  & 60.3  & 64.2   & -  & -  & 25.7   & 28.0 & - & 9.3                       & 7.5$^{\ddagger}$                 \\
HT-HAWP~\cite{lin2020deep}& 512                    & -  & \textcolor{blue}{62.9}  & \textcolor{blue}{66.6}   & - & -  & 25.0  & 27.4   & - & 10.5                       & 12.2$^{\ddagger}$                 \\
\midrule
L-CNN + \textit{M-LSD-s} & 512                    & 80.7  & 59.4  & 63.7   & 63.8 & 66.5  & \textcolor{Red}{27.5}  & 28.1   & 31.7 & 9.8                        & 16.6                 \\
HAWP + \textit{M-LSD-s} & 512                    & \textcolor{Red}{82.5}  & \textcolor{Red}{\textbf{63.3}}  & \textcolor{Red}{\textbf{67.1}}   & \textcolor{Red}{64.2} & \textcolor{Red}{66.7}  & \textcolor{Red}{27.5}  & \textcolor{Red}{\textbf{28.5}}   & \textcolor{Red}{32.4} & 10.4                       & 32.9                 \\
\textit{M-LSD-tiny}              & 320                    & 76.8  & 43.0  & 51.3   & 50.1 & 61.9  & 17.4  & 21.3   & 23.7 & \textcolor{Red}{\textbf{0.6}}                        & \textcolor{Red}{\textbf{200.8}}                \\
\textit{M-LSD-tiny}              & 512                    & 77.2  & 52.3  & 58.0   & 57.9 & 62.4  & 22.1  & 25.0   & 28.3 & \textcolor{Red}{\textbf{0.6}}                        & 164.1                \\
\textit{M-LSD}                   & 320                    & 78.7  & 48.2  & 55.5   & 55.7 & 63.4  & 20.2  & 23.9   & 27.7 & 1.5                        & 138.2                \\
\textit{M-LSD}                   & 512                    & \textcolor{Black}{80.0}  & \textcolor{Black}{56.4}  & \textcolor{Black}{62.1}   & \textcolor{Black}{61.5} & \textcolor{Black}{64.2}  & \textcolor{Black}{24.6}  & \textcolor{Black}{27.3}   & \textcolor{Black}{30.7} & 1.5                        & 115.4               \\
\bottomrule
\end{tabular}
\end{adjustbox}
\caption{Quantitative comparisons with existing LSD methods. FPS is evaluated in Tesla V100 GPU, where $^{\dagger}$ denotes CPU FPS and $^{\ddagger}$ denotes the values from the corresponding paper due to no published or incomplete implementation. $\ast$ denotes resizing the image with the shortest side at least 1100 pixels. M-LSD-s indicates the proposed training schemes. The best scores among previous methods, our models, and all together are marked in \textcolor{blue}{blue}, \textcolor{red}{red}, and \textbf{bold}, respectively.}
\label{table:sota}
\end{table*}

\subsection{Experimental Setting}
\textbf{Dataset and Evaluation Metrics. }
We evaluate our model with two famous LSD datasets: \textit{Wireframe}~\cite{huang2018learning} and \textit{YorkUrban}~\cite{denis2008efficient}.
The Wireframe dataset consists of 5,000 training and 462 test images of man-made environments, while the YorkUrban dataset has 102 test images.
Following the typical training and test protocol~\cite{huang2020tp,zhou2019end}, we train our model with the training set from the Wireframe dataset and test with both Wireframe and YorkUrban datasets.
We evaluate our models using prevalent metrics for LSD~\cite{huang2020tp,zhang2019ppgnet,meng2020lgnn,xue2019learning,zhou2019end} that include: heatmap-based metric $F^H$, structural average precision (sAP), and line matching average precision (LAP).

\textbf{Optimization. }
We train our model on Tesla V100 GPU.
We use the TensorFlow~\cite{abadi2016tensorflow} framework for model training and TFLite~\footnote{www.tensorflow.org/lite} for porting models to mobile devices.
Input images are resized to $320 \times 320$ or $512 \times 512$ in both training and testing, which are specified in each experiment.
The input augmentation consists of horizontal and vertical flips, shearing, rotation, and scaling.
We use ImageNet~\cite{deng2009imagenet} pre-trained weights on the parts of MobileNetV2~\cite{sandler2018mobilenetv2} in M-LSD and M-LSD-tiny.
Our model is trained using the Adam optimizer~\cite{kingma2014adam} with a learning rate of 0.01.
We use linear learning rate warm-up for 5 epochs and cosine learning rate decay~\cite{loshchilov2016sgdr} from 70 epoch to 150 epoch.
We train the model for a total of 150 epochs with a batch size of 64.

\subsection{Ablation Study and Interpretability}
\label{sec:ablation}
We conduct a series of ablation experiments to analyze our proposed method.
M-LSD-tiny is trained and tested on the Wireframe dataset with an input size of $512 \times 512$.
As shown in Table~\ref{table:ablation}, all the proposed schemes contribute to a significant performance improvement.
In addition, we include saliency map visualizations generated from each feature map to analyze networks learned from each training scheme in Figure~\ref{fig:comparison_cam} using GradCam~\cite{selvaraju2017grad}.
The saliency map interprets important regions and importance levels on the input image by computing the gradients from each feature map.

\textbf{Matching Loss. }
Integrating matching loss shows performance boosts on both pixel localization accuracy and line prediction quality.
We observe weak attention on center points from the baseline saliency maps in Figure~\ref{fig:match_x}, while w/ matching loss amplifies the attention on center points in Figure~\ref{fig:match_o}.
This demonstrates that training with coupled information of center points and displacement vectors allows the model to learn with more line-awareness features.

\textbf{Geometric Loss. }
Adding geometric loss gives performance boosts in every metric.
Moreover, the saliency map of Figure~\ref{fig:sol_x} shows more distinct and stronger attention on center points and line segments as compared to that of saliency maps w/ matching loss in Figure~\ref{fig:match_o}.
It shows that geometric information work as spatial attention cues for training.

\textbf{SoL Augmentation. }
Integrating SoL augmentation shows significant performance boost.
In the saliency maps of Figure~\ref{fig:sol_x}, w/ geometric loss shows strong but vague attention on center points with disconnected line attention for long line segments.
This can be a problem because the entire line information is essential to compute the center point.
In contrast, w/ SoL augmentation in Figure~\ref{fig:sol_o} shows more precise center point attention as well as clearly connected line attention.
This demonstrates that augmenting line segments by the number and length guides the model to be more robust in pixel-based and line matching-based qualities.

\subsection{Comparison with Other Methods}
\label{sec:sota}
As shown in Table~\ref{table:sota}, we conduct experiments that combine the proposed training schemes (SoL augmentation, matching and geometric loss) with existing methods.
Finally, we compare our proposed M-LSD and M-LSD-tiny with the previous state-of-the-art methods.

\textbf{Existing methods with M-LSD Training Schemes. }
As our proposed training schemes can be used with existing LSD methods, we demonstrate this using L-CNN and HAWP following Deep Hough Transform (HT)~\cite{lin2020deep}, a recently proposed combinable method.
L-CNN + HT (HT-L-CNN) shows a performance boost of 1.4\% while L-CNN + M-LSD-s shows a boost of 0.9\% in $sAP^{10}$.
HAWP + HT (HT-HAWP) shows 0.1\% of performance boost, while HAWP + M-LSD-s shows 0.6\% of performance boost in $sAP^{10}$, which makes the combination one of the state-of-the-art performance.
Thus, it demonstrates that the proposed training schemes are flexible and powerful to use with existing LSD methods.

\textbf{M-LSD and M-LSD-tiny. }
Our proposed models achieve competitive performance and the fastest inference speed even with a limited model size.
In comparison with the previous fastest model, TP-LSD-Lite, M-LSD with input size of 512 shows higher performance and an increase of 32.5\% in inference speed with only 6.3\% of the model size.
Our fastest model, M-LSD-tiny with 320 input size, has a slightly lower performance than that of TP-LSD-Lite, but achieves an increase of 130.5\% in inference speed with only 2.5\% of the model size.
Compared to the previous lightest model TP-LSD-HG, M-LSD with 512 input size outperforms on $sAP^{5}$, $sAP^{10}$ and $LAP$ with an increase of 136.0\% in inference speed with 20.3\% of the model size.
Our lightest model, M-LSD-tiny with 320 input size, shows an increase of 310.6\% in the inference speed with 8.1\% of the model size compared to TP-LSD-HG.
Previous methods can be deployed as real-time line segment detectors on server-class GPUs, but not on resource-constrained environments either because the model size is too large or the inference speed is too slow.
Although M-LSD does not achieve state-of-the-art performance, it shows competitive performance and the fastest inference speed with the smallest model size, offering the potential to be used in real-time applications on resource-constrained environments, such as mobile devices.

\begin{figure}[t!]
\centering
\includegraphics[width=1.0\columnwidth]{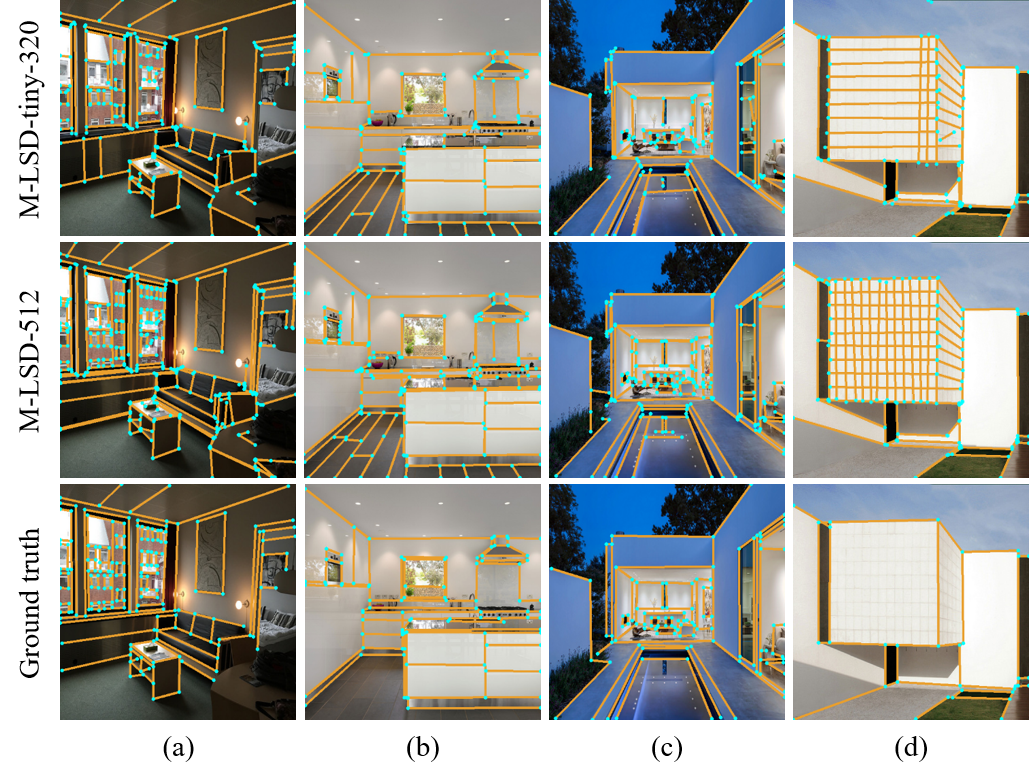}
\caption{Qualitative evaluation of M-LSD-tiny and M-LSD on WireFrame dataset.}
\label{fig:vis}
\end{figure}

\subsection{Visualization}
We visualize outputs of M-LSD and M-LSD-tiny in Figure~\ref{fig:vis}.
Junctions and line segments are colored with cyan blue and orange, respectively.
Compared to the GT, both models are capable of identifying junctions and line segments with high precision even in complicated low contrast environments such as (a) and (c).
Although the results of M-LSD-tiny may have a few small line segments missing and junctions incorrectly connected, the fundamental line segments to identify the environmental structure are accurate.

The goal of our model is to detect the structural line segments as~\cite{huang2018learning} while avoiding texture and photometric line segments.
However, we observe that some are included in our results, such as texture on the floor in (b) and shadow on the wall in (d).
We acknowledge this to be a common problem for existing methods, and considering texture and photometric features for training would be great future work.
We include more visualizations with a comparison of existing methods in the supplementary material.

\subsection{Deployment on Mobile Devices}
We deploy M-LSD on mobile devices and evaluate the memory usage and inference speed.
We use iPhone 12 Pro with A14 bionic chipset and Galaxy S20 Ultra with Snapdragon 865 ARM chipset.
As shown in Table~\ref{table:mobile}, M-LSD-tiny and M-LSD are small enough to be deployed on mobile devices where memory requirements range between 78MB and 508MB.
The inference speed of M-LSD-tiny is fast enough to be real-time on mobile devices where it ranges from a minimum of 17.9 FPS to a maximum of 56.8 FPS.
M-LSD still can be real-time with 320 input size, however, with 512 input size, FP16 may be required for a faster FPS over 10.
Overall, as all our models have small memory requirements and fast inference speed on mobile devices, the exceptional efficiency allows M-LSD variants to be used in real-world applications.
To the best of our knowledge, this is the first and the fastest real-time line segment detector on mobile devices ever reported.

\begin{table}[t!]
\centering
\begin{adjustbox}{width=1.0\columnwidth,center}
\begin{tabular}{lcccccc}
\toprule
Model                       & Input                & Device                   & FP & Latency (ms) & FPS  & Memory (MB) \\
\midrule
\multirow{8}{*}{M-LSD-tiny} & \multirow{4}{*}{320} & \multirow{2}{*}{iPhone}  & 32 & 30.6         & 32.7 & 169         \\
                            &                      &                          & 16 & 20.6         & 48.6 & 111         \\ \cline{3-7} 
                            &                      & \multirow{2}{*}{Android} & 32 & 31.0         & 32.3 & 103         \\
                            &                      &                          & 16 & \textbf{17.6}         & \textbf{56.8} & \textbf{78}          \\ \cline{2-7} 
                            & \multirow{4}{*}{512} & \multirow{2}{*}{iPhone}  & 32 & 51.6         & 19.4 & 203         \\
                            &                      &                          & 16 & 36.8         & 27.1 & 176         \\ \cline{3-7} 
                            &                      & \multirow{2}{*}{Android} & 32 & 55.8         & 17.9 & 195         \\
                            &                      &                          & 16 & \textbf{25.4}         & \textbf{39.4} & \textbf{129}         \\
\midrule
\multirow{8}{*}{M-LSD}      & \multirow{4}{*}{320} & \multirow{2}{*}{iPhone}  & 32 & 74.5         & 13.4 & 241         \\
                            &                      &                          & 16 & 46.4         & 21.6 & 188         \\ \cline{3-7} 
                            &                      & \multirow{2}{*}{Android} & 32 & 82.4         & 12.1 & 236         \\
                            &                      &                          & 16 & \textbf{38.4}         & \textbf{26.0} & \textbf{152}         \\ \cline{2-7} 
                            & \multirow{4}{*}{512} & \multirow{2}{*}{iPhone}  & 32 & 121.6        & 8.2  & 327         \\
                            &                      &                          & 16 & 90.7         & 11.0 & \textbf{261}         \\ \cline{3-7} 
                            &                      & \multirow{2}{*}{Android} & 32 & 177.3        & 5.6  & 508         \\
                            &                      &                          & 16 & \textbf{79.0}         & \textbf{12.7} & 289         \\
\bottomrule
\end{tabular}
\end{adjustbox}
\caption{Inference speed and memory usage on iPhone (A14 Bionic chipset) and Android phone (Snapdragon 865 chipset). FP denotes floating point.}
\label{table:mobile}
\end{table}

\section{Conclusion}
We introduce M-LSD, a light-weight and real-time line segment detector for resource-constrained environments.
Our model is designed with a significantly efficient network architecture and a single module process to predict line segments.
To maintain competitive performance even with a light-weight network, we present novel training schemes: SoL augmentation, matching and geometric loss.
As a result, our proposed method achieves competitive performance and the fastest inference speed with the lightest model size.
Moreover, we show that M-LSD is deployable on mobile devices in real-time, which demonstrates the potential to be used in real-time mobile applications.
     
\bibliography{aaai22}

\begin{thebibliography}{36}
\providecommand{\natexlab}[1]{#1}

\bibitem[{Abadi et~al.(2016)Abadi, Agarwal, Barham, Brevdo, Chen, Citro,
  Corrado, Davis, Dean, Devin et~al.}]{abadi2016tensorflow}
Abadi, M.; Agarwal, A.; Barham, P.; Brevdo, E.; Chen, Z.; Citro, C.; Corrado,
  G.~S.; Davis, A.; Dean, J.; Devin, M.; et~al. 2016.
\newblock Tensorflow: Large-scale machine learning on heterogeneous distributed
  systems.
\newblock \emph{arXiv preprint arXiv:1603.04467}.

\bibitem[{Bartoli and Sturm(2005)}]{bartoli2005structure}
Bartoli, A.; and Sturm, P. 2005.
\newblock Structure-from-motion using lines: Representation, triangulation, and
  bundle adjustment.
\newblock \emph{Computer vision and image understanding}, 100(3): 416--441.

\bibitem[{Deng et~al.(2009)Deng, Dong, Socher, Li, Li, and
  Fei-Fei}]{deng2009imagenet}
Deng, J.; Dong, W.; Socher, R.; Li, L.-J.; Li, K.; and Fei-Fei, L. 2009.
\newblock Imagenet: A large-scale hierarchical image database.
\newblock In \emph{2009 IEEE conference on computer vision and pattern
  recognition}, 248--255. Ieee.

\bibitem[{Denis, Elder, and Estrada(2008)}]{denis2008efficient}
Denis, P.; Elder, J.~H.; and Estrada, F.~J. 2008.
\newblock Efficient edge-based methods for estimating manhattan frames in urban
  imagery.
\newblock In \emph{European conference on computer vision}, 197--210. Springer.

\bibitem[{Faugeras et~al.(1992)Faugeras, Deriche, Mathieu, Ayache, and
  Randall}]{faugeras1992depth}
Faugeras, O.~D.; Deriche, R.; Mathieu, H.; Ayache, N.; and Randall, G. 1992.
\newblock The depth and motion analysis machine.
\newblock In \emph{Parallel Image Processing}, 143--175. World Scientific.

\bibitem[{Girshick(2015)}]{girshick2015fast}
Girshick, R. 2015.
\newblock Fast r-cnn.
\newblock In \emph{Proceedings of the IEEE international conference on computer
  vision}, 1440--1448.

\bibitem[{Girshick et~al.(2014)Girshick, Donahue, Darrell, and
  Malik}]{girshick2014rcnn}
Girshick, R.; Donahue, J.; Darrell, T.; and Malik, J. 2014.
\newblock Rich feature hierarchies for accurate object detection and semantic
  segmentation.
\newblock In \emph{Proceedings of the IEEE conference on computer vision and
  pattern recognition}, 580--587.

\bibitem[{Howard et~al.(2017)Howard, Zhu, Chen, Kalenichenko, Wang, Weyand,
  Andreetto, and Adam}]{howard2017mobilenets}
Howard, A.~G.; Zhu, M.; Chen, B.; Kalenichenko, D.; Wang, W.; Weyand, T.;
  Andreetto, M.; and Adam, H. 2017.
\newblock Mobilenets: Efficient convolutional neural networks for mobile vision
  applications.
\newblock \emph{arXiv preprint arXiv:1704.04861}.

\bibitem[{Huang and Gao(2019)}]{huang2019wireframe}
Huang, K.; and Gao, S. 2019.
\newblock Wireframe parsing with guidance of distance map.
\newblock \emph{IEEE Access}, 7: 141036--141044.

\bibitem[{Huang et~al.(2018)Huang, Wang, Zhou, Ding, Gao, and
  Ma}]{huang2018learning}
Huang, K.; Wang, Y.; Zhou, Z.; Ding, T.; Gao, S.; and Ma, Y. 2018.
\newblock Learning to parse wireframes in images of man-made environments.
\newblock In \emph{Proceedings of the IEEE Conference on Computer Vision and
  Pattern Recognition}, 626--635.

\bibitem[{Huang et~al.(2020)Huang, Qin, Xiong, Ding, He, and Liu}]{huang2020tp}
Huang, S.; Qin, F.; Xiong, P.; Ding, N.; He, Y.; and Liu, X. 2020.
\newblock TP-LSD: Tri-Points Based Line Segment Detector.
\newblock \emph{arXiv preprint arXiv:2009.05505}.

\bibitem[{Kingma and Ba(2014)}]{kingma2014adam}
Kingma, D.~P.; and Ba, J. 2014.
\newblock Adam: A method for stochastic optimization.
\newblock \emph{arXiv preprint arXiv:1412.6980}.

\bibitem[{Li et~al.(2018)Li, Li, Lin, and Li}]{li2018tiny}
Li, Y.; Li, J.; Lin, W.; and Li, J. 2018.
\newblock Tiny-DSOD: Lightweight object detection for resource-restricted
  usages.
\newblock \emph{arXiv preprint arXiv:1807.11013}.

\bibitem[{Lin, Pintea, and van Gemert(2020)}]{lin2020deep}
Lin, Y.; Pintea, S.~L.; and van Gemert, J.~C. 2020.
\newblock Deep hough-transform line priors.
\newblock In \emph{European Conference on Computer Vision}, 323--340. Springer.

\bibitem[{Liu et~al.(2016)Liu, Anguelov, Erhan, Szegedy, Reed, Fu, and
  Berg}]{liu2016ssd}
Liu, W.; Anguelov, D.; Erhan, D.; Szegedy, C.; Reed, S.; Fu, C.-Y.; and Berg,
  A.~C. 2016.
\newblock Ssd: Single shot multibox detector.
\newblock In \emph{European conference on computer vision}, 21--37. Springer.

\bibitem[{Loshchilov and Hutter(2016)}]{loshchilov2016sgdr}
Loshchilov, I.; and Hutter, F. 2016.
\newblock Sgdr: Stochastic gradient descent with warm restarts.
\newblock \emph{arXiv preprint arXiv:1608.03983}.

\bibitem[{Meng et~al.(2020)Meng, Zhang, Hu, He, and Yu}]{meng2020lgnn}
Meng, Q.; Zhang, J.; Hu, Q.; He, X.; and Yu, J. 2020.
\newblock LGNN: A Context-aware Line Segment Detector.
\newblock In \emph{Proceedings of the 28th ACM International Conference on
  Multimedia}, 4364--4372.

\bibitem[{Micusik and Wildenauer(2017)}]{micusik2017structure}
Micusik, B.; and Wildenauer, H. 2017.
\newblock Structure from motion with line segments under relaxed endpoint
  constraints.
\newblock \emph{International Journal of Computer Vision}, 124(1): 65--79.

\bibitem[{P{\v{r}}ibyl, Zem{\v{c}}{\'\i}k, and
  {\v{C}}ad{\'\i}k(2017)}]{pvribyl2017absolute}
P{\v{r}}ibyl, B.; Zem{\v{c}}{\'\i}k, P.; and {\v{C}}ad{\'\i}k, M. 2017.
\newblock Absolute pose estimation from line correspondences using direct
  linear transformation.
\newblock \emph{Computer Vision and Image Understanding}, 161: 130--144.

\bibitem[{Redmon et~al.(2016)Redmon, Divvala, Girshick, and
  Farhadi}]{redmon2016you}
Redmon, J.; Divvala, S.; Girshick, R.; and Farhadi, A. 2016.
\newblock You only look once: Unified, real-time object detection.
\newblock In \emph{Proceedings of the IEEE conference on computer vision and
  pattern recognition}, 779--788.

\bibitem[{Redmon and Farhadi(2017)}]{redmon2017yolo9000}
Redmon, J.; and Farhadi, A. 2017.
\newblock YOLO9000: better, faster, stronger.
\newblock In \emph{Proceedings of the IEEE conference on computer vision and
  pattern recognition}, 7263--7271.

\bibitem[{Redmon and Farhadi(2018)}]{redmon2018yolov3}
Redmon, J.; and Farhadi, A. 2018.
\newblock Yolov3: An incremental improvement.
\newblock \emph{arXiv preprint arXiv:1804.02767}.

\bibitem[{Ren et~al.(2015)Ren, He, Girshick, and Sun}]{ren2015faster}
Ren, S.; He, K.; Girshick, R.; and Sun, J. 2015.
\newblock Faster r-cnn: Towards real-time object detection with region proposal
  networks.
\newblock \emph{arXiv preprint arXiv:1506.01497}.

\bibitem[{Sandler et~al.(2018)Sandler, Howard, Zhu, Zhmoginov, and
  Chen}]{sandler2018mobilenetv2}
Sandler, M.; Howard, A.; Zhu, M.; Zhmoginov, A.; and Chen, L.-C. 2018.
\newblock Mobilenetv2: Inverted residuals and linear bottlenecks.
\newblock In \emph{Proceedings of the IEEE conference on computer vision and
  pattern recognition}, 4510--4520.

\bibitem[{Selvaraju et~al.(2017)Selvaraju, Cogswell, Das, Vedantam, Parikh, and
  Batra}]{selvaraju2017grad}
Selvaraju, R.~R.; Cogswell, M.; Das, A.; Vedantam, R.; Parikh, D.; and Batra,
  D. 2017.
\newblock Grad-cam: Visual explanations from deep networks via gradient-based
  localization.
\newblock In \emph{Proceedings of the IEEE international conference on computer
  vision}, 618--626.

\bibitem[{Von~Gioi et~al.(2008)Von~Gioi, Jakubowicz, Morel, and
  Randall}]{von2008lsd}
Von~Gioi, R.~G.; Jakubowicz, J.; Morel, J.-M.; and Randall, G. 2008.
\newblock LSD: A fast line segment detector with a false detection control.
\newblock \emph{IEEE transactions on pattern analysis and machine
  intelligence}, 32(4): 722--732.

\bibitem[{Wang, Li, and Ling(2018)}]{wang2018pelee}
Wang, R.~J.; Li, X.; and Ling, C.~X. 2018.
\newblock Pelee: A real-time object detection system on mobile devices.
\newblock \emph{arXiv preprint arXiv:1804.06882}.

\bibitem[{Xu et~al.(2016)Xu, Zhang, Cheng, and Koch}]{xu2016pose}
Xu, C.; Zhang, L.; Cheng, L.; and Koch, R. 2016.
\newblock Pose estimation from line correspondences: A complete analysis and a
  series of solutions.
\newblock \emph{IEEE transactions on pattern analysis and machine
  intelligence}, 39(6): 1209--1222.

\bibitem[{Xu et~al.(2021)Xu, Xu, Cheung, and Tu}]{xu2021line}
Xu, Y.; Xu, W.; Cheung, D.; and Tu, Z. 2021.
\newblock Line segment detection using transformers without edges.
\newblock In \emph{Proceedings of the IEEE/CVF Conference on Computer Vision
  and Pattern Recognition}, 4257--4266.

\bibitem[{Xue et~al.(2019{\natexlab{a}})Xue, Bai, Wang, Xia, Wu, and
  Zhang}]{xue2019learning}
Xue, N.; Bai, S.; Wang, F.; Xia, G.-S.; Wu, T.; and Zhang, L.
  2019{\natexlab{a}}.
\newblock Learning attraction field representation for robust line segment
  detection.
\newblock In \emph{Proceedings of the IEEE/CVF Conference on Computer Vision
  and Pattern Recognition}, 1595--1603.

\bibitem[{Xue et~al.(2020)Xue, Wu, Bai, Wang, Xia, Zhang, and
  Torr}]{xue2020holistically}
Xue, N.; Wu, T.; Bai, S.; Wang, F.; Xia, G.-S.; Zhang, L.; and Torr, P.~H.
  2020.
\newblock Holistically-attracted wireframe parsing.
\newblock In \emph{Proceedings of the IEEE/CVF Conference on Computer Vision
  and Pattern Recognition}, 2788--2797.

\bibitem[{Xue et~al.(2017)Xue, Xia, Bai, Zhang, and Shen}]{xue2017anisotropic}
Xue, N.; Xia, G.-S.; Bai, X.; Zhang, L.; and Shen, W. 2017.
\newblock Anisotropic-scale junction detection and matching for indoor images.
\newblock \emph{IEEE Transactions on Image Processing}, 27(1): 78--91.

\bibitem[{Xue, Zhou, and Huang(2019)}]{xue2019neural}
Xue, Y.; Zhou, Z.; and Huang, X. 2019.
\newblock Neural Wireframe Renderer: Learning Wireframe to Image Translations.
\newblock \emph{arXiv preprint arXiv:1912.03840}.

\bibitem[{Xue et~al.(2019{\natexlab{b}})Xue, Xue, Xia, and
  Shen}]{xue2019rectification}
Xue, Z.; Xue, N.; Xia, G.-S.; and Shen, W. 2019{\natexlab{b}}.
\newblock Learning to calibrate straight lines for fisheye image rectification.
\newblock In \emph{Proceedings of the IEEE/CVF Conference on Computer Vision
  and Pattern Recognition}, 1643--1651.

\bibitem[{Zhang et~al.(2019)Zhang, Li, Bi, Zheng, Wang, Huang, Luo, Xu, and
  Gao}]{zhang2019ppgnet}
Zhang, Z.; Li, Z.; Bi, N.; Zheng, J.; Wang, J.; Huang, K.; Luo, W.; Xu, Y.; and
  Gao, S. 2019.
\newblock Ppgnet: Learning point-pair graph for line segment detection.
\newblock In \emph{Proceedings of the IEEE/CVF Conference on Computer Vision
  and Pattern Recognition}, 7105--7114.

\bibitem[{Zhou, Qi, and Ma(2019)}]{zhou2019end}
Zhou, Y.; Qi, H.; and Ma, Y. 2019.
\newblock End-to-end wireframe parsing.
\newblock In \emph{Proceedings of the IEEE/CVF International Conference on
  Computer Vision}, 962--971.

\end{thebibliography}

\clearpage



\section*{Supplementary material}

# Towards Light-weight and Real-time Line Segment Detection

*Supplementary Material*

github.com/navervision/mlsd

# Contents

| Block | Input | SC input | Operator | c | n |
|---|---|---|---|---|---|
| 1 | H×W×3 | - | conv2d | 32 | 1 |
| 2 | H/2×W/2×32 | - | bottleneck | 16 | 1 |
| 3∼4 | H/2×W/2×16 | - | bottleneck | 24 | 2 |
| 5∼7 | H/4×W/4×24 | - | bottleneck | 32 | 3 |
| 8∼11 | H/8×W/8×32 | - | bottleneck | 64 | 4 |
| 12∼14 | H/16×W/16×64 | - | bottleneck | 96 | 3 |
| 15 | H/16×W/16×96 | H/16×W/16×64 | block type A | 128 | 1 |
| 16 | H/16×W/16×128 | - | block type B | 64 | 1 |
| 17 | H/16×W/16×64 | H/8×W/8×32 | block type A | 128 | 1 |
| 18 | H/8×W/8×128 | - | block type B | 64 | 1 |
| 19 | H/8×W/8×64 | H/4×W/4×24 | block type A | 128 | 1 |
| 20 | H/4×W/4×128 | - | block type B | 64 | 1 |
| 21 | H/4×W/4×64 | H/2×W/2×16 | block type A | 128 | 1 |
| 22 | H/2×W/2×128 | - | block type B | 64 | 1 |
| 23 | H/2×W/2×64 | - | block type C | 16 | 1 |
| Final | H/2×W/2×16 | - | - | - | - |

(a) M-LSD

| Block | Input | SC input | Operator | c | n |
|---|---|---|---|---|---|
| 1 | H×W×3 | - | conv2d | 32 | 1 |
| 2 | H/2×W/2×32 | - | bottleneck | 16 | 1 |
| 3∼4 | H/2×W/2×16 | - | bottleneck | 24 | 2 |
| 5∼7 | H/4×W/4×24 | - | bottleneck | 32 | 3 |
| 8∼11 | H/8×W/8×32 | - | bottleneck | 64 | 4 |
| 12 | H/16×W/16×64 | H/8×W/8×32 | block type A | 128 | 1 |
| 13 | H/8×W/8×128 | - | block type B | 64 | 1 |
| 14 | H/8×W/8×64 | H/4×W/4×24 | block type A | 64 | 1 |
| 15 | H/4×W/4×64 | - | block type B | 64 | 1 |
| 16 | H/4×W/4×64 | - | block type C | 16 | 1 |
| - | H/4×W/4×16 | - | upscale | 16 | 1 |
| Final | H/2×W/2×16 | - | - | - | - |

(b) M-LSD-tiny

Table A: Architecture details of M-LSD and M-LSD-tiny. Each line describes a sequence of 1 or repeating $n$ identical layers where each layer in the same sequence has the same $c$ output channels. Block numbers ('Block') and block type A∼C in 'Operator' are from Figure 3 and Figure A. 'SC input' denotes a skip connection input and the bottleneck operation is from MobileNetV2 (Sandler et al. 2018).

# A  Additional Related Works

**Hand-crafted Feature-based Methods.** For a long period of time, hand-crafted low-level features, especially line gradients, have been used for LSD. These conventional approaches can be categorized into edge map based and perceptual grouping methods. Edge map based methods (Kamat-Sadekar and Ganesan 1998; Furukawa and Shinagawa 2003; Matas, Galambos, and Kittler 2000; Xu, Shin, and Klette 2014) convert the pixel-wise feature map of an image to a parameter map by Hough transform to sort out line predictions. A key challenge of these methods is to identify endpoints of the line segment (Elder et al. 2017). Perceptual grouping methods (Von Gioi et al. 2008; Cho, Yuille, and Lee 2017; Burns, Hanson, and Riseman 1986) exploit the image gradients as geometry cues to group pixels into line segment candidates. However, choosing an appropriate threshold to discriminate true line segments remains a challenge in these methods. In (Almazan et al. 2017), there has been an attempt to merge both approaches. The method first uses the probabilistic Hough method to identify optimal lines; then, localize the line segments that generated the peak in the Hough map.

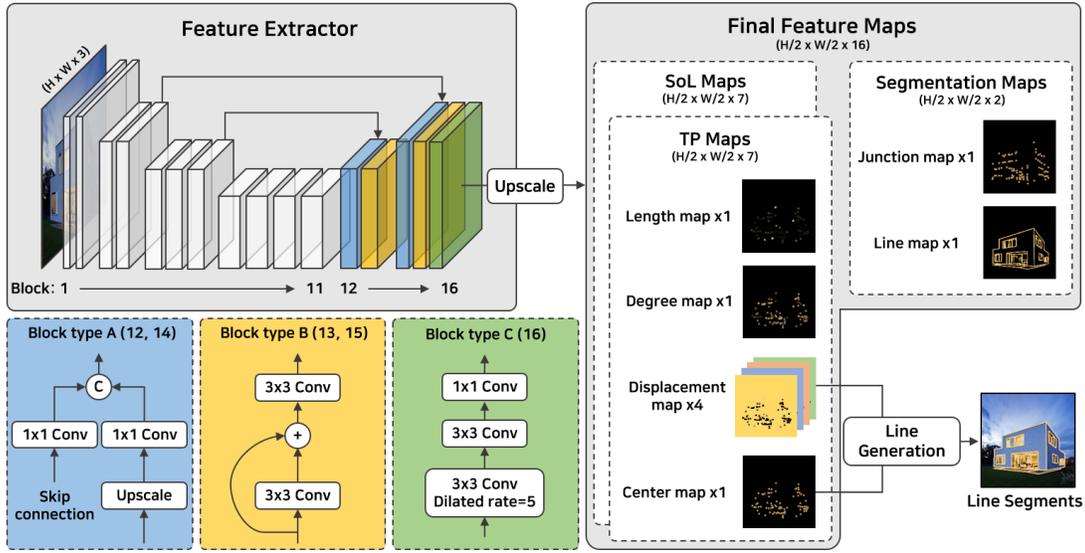

Figure A: The overall architecture of M-LSD-tiny. In the feature extractor, block 1 ∼ 11 are parts of MobileNetV2, and block 12 ∼ 16 are designed as a top-down architecture. The final feature maps are simply generated by upscale. The predicted line segments are generated by merging center points and displacement vectors from the TP maps.

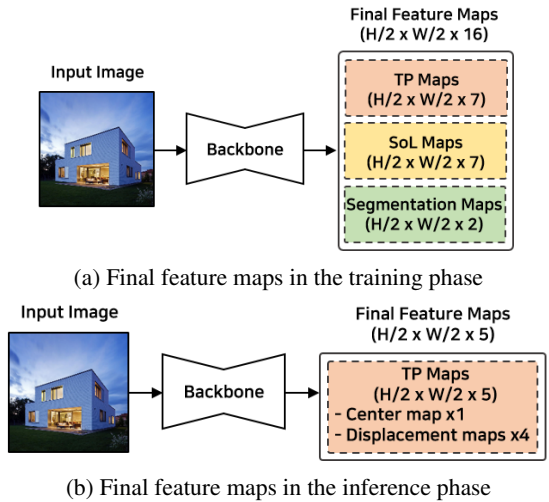

(a) Final feature maps in the training phase

(b) Final feature maps in the inference phase

Figure B: Final feature maps in the training and inference phase. (a) In the training phase, the final feature maps include TP, SoL, and segmentation maps with a total of 16 channels. (b) For better efficiency in the inference phase, we disregard unnecessary convolutions and maintain only the center and displacement maps in the TP maps with a total of 5 channels.

# B  Details of M-LSD

## B.1  Network Architecture

The detailed architecture of M-LSD and M-LSD-tiny is described in Table A. M-LSD includes an encoder structure from MobileNetV2 (Sandler et al. 2018) in block 1∼14 and designed decoder structure in block 15∼final. M-LSD-tiny also includes an encoder structure from MobileNetV2 in block 1∼11 and a custom decoder structure in block 12∼final, which is illustrated in Figure A. The final feature maps in M-LSD-tiny are generated by upscaling with $H/2 \times W/2 \times 16$ tensors when the input image is $H \times W \times 3$. For the upscale operation, we use bilinear interpolation. On the other hand, M-LSD uses the feature map from block type C as a final feature map with the same size of $H/2 \times W/2 \times 16$.

## B.2  Feature Maps and Losses

In (Huang et al. 2020), the weighted binary cross-entropy (WBCE) loss is used to train the center map. However, we observe that the number of positive (foreground) pixels is much less than that of negative (background) pixels, and such foreground-background class imbalance degrades the performance of the WBCE loss. This is because the majority of pixels are easy negatives that contribute no useful learning signals. Thus, we separate positive and negative terms of the binary cross-entropy loss to have the same scale, and reformulate a separate binary classification loss as follows:

$$\ell_{pos}(F) = \frac{-1}{\sum_p I(p)} \sum_p W(p) \cdot log\sigma(F(p)), \quad \text{(i)}$$

$$\ell_{neg}(F) = \frac{-1}{\sum_p 1-I(p)} \sum_p (1 - I(p)) \cdot log(1 - \sigma(F(p))), \quad \text{(ii)}$$

$$\ell_{cls}(F) = \lambda_{pos} \cdot \ell_{pos}(F) + \lambda_{neg} \cdot \ell_{neg}(F), \quad \text{(iii)}$$

where $I(p)$ outputs 1 if the pixel $p$ of the GT map is non-zero, otherwise 0, $\sigma$ denotes a sigmoid function, and $W(p)$ and $F(p)$ are pixel values in the GT and feature map, respectively. For the GT of the center map, positions of the center point are marked on a zero map, which is then scaled using a Gaussian kernel with 5 stdev, truncated by a $3 \times 3$ window. We use the center loss as $\mathcal{L}_{center} = \ell_{cls}(C)$, where $C$ denotes the center map and weights ($\lambda_{pos}, \lambda_{neg}$) set to (1,30).

For the displacement maps, we compute displacement vectors from the ground truth (GT) and mark those values on

| Model | Parts of MNV2 in encoder | Params (M) | | | Inference speed (FPS) | | | Performance | | |
|---|---|---|---|---|---|---|---|---|---|---|
| | | Encoder (% of MNV2) | Decoder | Total | Backbone | Prediction | Total | $F^H$ | $sAP^{10}$ | LAP |
| M-LSD-tiny | Input $\sim$ 64-channel | **0.3 (7.4)** | **0.3** | **0.6** | **201.6** | 881.9 | **164.1** | 77.2 | 58.0 | 57.9 |
| M-LSD | Input $\sim$ 96-channel | 0.6 (16.5) | 0.9 | 1.5 | 132.8 | 883.4 | 115.4 | 80.0 | 62.1 | 61.5 |
| 1 | Input $\sim$ 160-channel | 1.0 (30.6) | 1.3 | 2.3 | 124.7 | 885.1 | 109.3 | 79.9 | **62.8** | 62.4 |
| 2 | Input $\sim$ 320-channel | 1.8 (54.1) | 1.5 | 3.3 | 117.9 | **885.7** | 104.0 | 79.7 | 62.5 | **62.6** |
| 3 | Input $\sim$ 1280-channel | 2.3 (66.5) | 1.7 | 4.0 | 107.6 | 883.4 | 95.9 | **80.2** | **62.8** | 62.1 |

(a) Ablation study by varying the parts used from the MobileNetV2 (MNV2) for the encoder architecture. Performance is reported on Wireframe dataset. '% of MNV2' indicates the percentage of parameters used in each type of encoder compared to the total parameters used in MobileNetV2.

| Model | Setup | Params (M) | Inference speed (FPS) | | | Performance | | |
|---|---|---|---|---|---|---|---|---|
| | | | Backbone | Prediction | Total | $F^H$ | $sAP^{10}$ | LAP |
| M-LSD-tiny | Block type A: $1 \times 1$ conv / B: pre-residual / C: dilated rate 5 | **0.6** | 201.6 | 881.9 | 164.1 | **77.2** | 58.0 | 57.9 |
| 4 | Block type A: $1 \times 1$ conv $\to 3 \times 3$ conv | 0.7 | 199.2 | 881.9 | 162.5 | 76.7 | **58.1** | 57.9 |
| 5 | Block type B: pre-residual $\to$ post-residual | 0.7 | 200.5 | 881.9 | 163.4 | 76.9 | **58.1** | 58.0 |
| 6 | Block type C: dilated rate $5 \to 1$ | **0.6** | **215.2** | 881.9 | **173.0** | 75.9 | 56.1 | 56.0 |
| 7 | Block type C: dilated rate $5 \to 3$ | **0.6** | 203.5 | 881.9 | 165.3 | 76.7 | 57.6 | 57.4 |

(b) Ablation study by varying block types for the decoder architecture. Performance is reported on Wireframe dataset with M-LSD-tiny as the baseline. Block type A $\sim$ B are from Figure 3 and Figure A.

Table B: Ablation study on encoder and decoder architectures.

the center of line segment in the GT map. Next, these values are extrapolated to a 3×3 window (center blob) so that all neighboring pixels of a given pixel contain the same value. For the displacement, length, and degree maps, we use the smooth L1 loss for regression learning. The regression loss can be formulated as follows:

$$\ell_{reg}(F) = \frac{1}{\sum_p H(p)} \sum_p H(p) \cdot L_1^{smooth}(F(p), \hat{F}(p)), \quad \text{(iv)}$$

where $F(p)$ and $\hat{F}(p)$ denote values of pixel $p$ in the feature map $F$ and the GT map $\hat{F}$, and $H(p)$ outputs 1 if the pixel $p$ of the GT map is on the center blob (extrapolated $3 \times 3$ window). We use the displacement loss $\mathcal{L}_{disp} = \ell_{reg}(D)$, where $D$ denotes the displacement map. The length and degree losses are $\mathcal{L}_{length} = \ell_{reg}(\sigma(L))$ and $\mathcal{L}_{degree} = \ell_{reg}(\sigma(G))$, where $\sigma(L)$ and $\sigma(G)$ are sigmoid functions $\sigma$ applied to length and degree maps. Note that only the GT points and its neighboring pixels in $3 \times 3$ window are used for the loss computation. In the line generation process, the center map is applied with a sigmoid function to output a probability value, while the displacement map uses the original values. Then, we extract the exact center point position by non-maximum suppression (Huang et al. 2018; Zhou, Qi, and Ma 2019; Huang et al. 2020) on the center map to remove duplicates around correct predictions.

### B.3 Usage of Final Feature Maps

In the training phase, M-LSD and M-LSD-tiny outputs final feature maps of 16 channels, which include 7 channels for TP maps, 7 channels for SoL maps, and 2 channels for segmentation maps as illustrated in Figure Ba. However, as the line generation process only requires the center and displacement maps of TP maps, operations for the other auxiliary maps are unnecessary in the inference phase. Thus, we disregard these operations and output only 5 channels of TP maps in the inference phase, including 1 center map and 4 displacement maps, as shown in Figure Bb. As a result, we can minimize computational cost and maximize the inference speed.

## C Extended Experiments
### C.1 Ablation Study of Architecture

We run a series of ablation experiments to investigate various encoder and decoder architectures. As shown in Table Ba, we vary the parts used from the MobileNetV2 on the encoder architecture. As the encoder size increases, we add block types A and B to the decoder structure by following the structural format in Table Aa. Model 1 $\sim$ 3 exploit bigger and deeper encoder architectures, which result in larger model parameters and slower inference speed. The performance turns out to be slightly higher than that of M-LSD. However, we choose 'Input $\sim$ 96-channel' of MobileNetV2 as the encoder for M-LSD because increasing the encoder size causes larger amounts of model parameters to be used and decreases the inference speed with a negligible performance boost. Therefore, we observe that 'Input $\sim$ 96-channel' is the largest model that can run on a mobile device in real-time. In contrast, when performing real-time LSD on GPUs, model 1 $\sim$ 3 are good candidates as they outperform TP-LSD-Lite (Huang et al. 2020), previously the best real-time LSD, with faster inference speed and lighter model size.

In Table Bb, we vary the block types used in the decoder architecture. Model 4 changes every $1 \times 1$ convolution to a $3 \times 3$ convolution in block type A, while model 5 changes the residual connection from being in between the convolutions ('pre-residual') to the end of the convolutions ('post-residual') for block type B. These changes result in an in-

| Setup | Params | Inference speed (FPS) | | | Performance | | |
|---|---|---|---|---|---|---|---|
| | | Backbone | Prediction | Total | $F^H$ | $sAP^{10}$ | LAP |
| w/o offset | 629253 | 201.6 | **881.9** | **164.1** | 77.2 | **58.0** | 57.9 |
| w/ offset | 629383 | 201.6 | 811.4 | 161.5 | 77.2 | 57.9 | 57.9 |

Table C: Experiments of w/o and w/ offset maps in M-LSD-tiny on Wireframe dataset.

| $\epsilon$ | $\mu$ | # origin | # aug | # total | $F^H$ | $sAP^{10}$ | LAP |
|---|---|---|---|---|---|---|---|
| 0.000 | - | 374884 | 0 | 374884 | 76.2 | 55.1 | 55.3 |
| 0.050 | 25.6 | 374884 | 851555 | 1226439 | 76.2 | 56.2 | 56.3 |
| 0.100 | 51.2 | 374884 | 251952 | 626836 | 76.4 | 57.2 | 57.3 |
| 0.125 | 64.0 | 374884 | 151804 | 526688 | **77.2** | **58.0** | **57.9** |
| 0.150 | 76.8 | 374884 | 102719 | 477603 | 77.0 | 57.5 | 57.9 |
| 0.200 | 102.4 | 374884 | 47500 | 422384 | 76.6 | 56.8 | 56.5 |
| 0.300 | 153.6 | 374884 | 12123 | 387007 | 76.6 | 56.1 | 56.7 |
| 0.400 | 204.8 | 374884 | 3250 | 378134 | 76.4 | 55.5 | 56.1 |
| 0.500 | 256.0 | 374884 | 170 | 375054 | 76.2 | 55.0 | 55.7 |

Table D: Impact of ratio $\epsilon$ in SoL augmentation with M-LSD-tiny on Wireframe dataset. $\epsilon = 0.0$ is the baseline with no SoL augmentation applied. The base length of subpart $\mu$ is computed by $\mu =$ input size $\times \epsilon$. '# origin', '# aug', and '# total' denote the number of original, augmented, and total line segments.

| | # origin | # aug | # total | $F^H$ | $sAP^{10}$ | LAP |
|---|---|---|---|---|---|---|
| baseline | 374884 | - | 374884 | 76.2 | 55.1 | 55.3 |
| w/ overlap | 374884 | 151804 | 526688 | **77.2** | **58.0** | **57.9** |
| w/o overlap | 374884 | 41101 | 415985 | 76.4 | 56.7 | 56.7 |

Table E: Impact of overlapping in SoL augmentation with M-LSD-tiny on Wireframe dataset. The baseline is not trained with SoL augmentation. '# origin', '# aug', and '# total' denote the number of original, augmented, and total line segments.

| $\gamma$ | Input size 320 | | | Input size 512 | | |
|---|---|---|---|---|---|---|
| | $F^H$ | $sAP^{10}$ | LAP | $F^H$ | $sAP^{10}$ | LAP |
| 0.0 | 75.9 | 47.1 | 44.9 | 76.1 | 55.1 | 54.8 |
| 2.5 | 76.2 | 50.4 | 48.9 | 76.5 | 57.2 | 57.2 |
| 5.0 | **76.8** | **51.3** | **50.1** | 77.2 | 58.0 | **57.9** |
| 7.5 | 76.0 | 49.0 | 48.5 | 76.8 | **58.5** | 57.2 |
| 10.0 | 75.0 | 45.1 | 45.0 | 76.8 | 57.8 | 56.7 |
| 12.5 | 74.1 | 43.1 | 43.2 | 76.2 | 56.7 | 55.8 |
| 15.0 | 74.2 | 42.7 | 42.8 | 75.7 | 54.0 | 53.2 |
| 20.0 | 73.6 | 41.4 | 42.1 | 75.1 | 51.0 | 50.6 |

Table F: Impact of matching loss threshold $\gamma$ with M-LSD-tiny on Wireframe dataset. $\gamma = 0.0$ is the baseline with no matching loss applied.

crease in model size and a decrease in inference speed because 'post-residual' requires twice the number of output channels than that of 'pre-residual'. However, the performance remains similar to that of M-LSD-tiny. For models 6 and 7, the dilated rate of the first convolution in block type C is changed to 1 and 3, respectively. Here we observe that by decreasing the dilated rate can improve the inference speed but conversely decrease the performance. This is because the dilated convolution can effectively manage long line segments, which require large receptive fields. Thus, we choose to use $1 \times 1$ convolution in block type A, 'pre-residual' in block type B, and the dilated rate of 5 in block type C.

### C.2 Needs of Offset Maps

In some of the previous LSD methods (Meng et al. 2020; Zhou, Qi, and Ma 2019; Xue et al. 2020), offset maps are used to estimate offsets between the predicted map and input image because the predicted map has a smaller resolution than the input image. We perform experiments and evaluate the effectiveness of offset maps with M-LSD-tiny. When we apply offset maps to M-LSD-tiny, we need two offset maps for the center point (one for each coordinate). As shown in Table C, w/ offset maps increase in model parameters and decrease in inference speed, while the performance does not change. This demonstrates that offset maps are unnecessary for M-LSD-tiny because the resolution of the input image is two times the size of the resolution of predicted maps, which is minor. Thus, we disregard offset maps in M-LSD architectures.

### C.3 Impact of SoL Augmentation

In SoL augmentation, the number of internally dividing points $k$ is based on the length of the line segment and computed as $k = \lfloor r(l)/(\mu/2) \rfloor - 1$, where $r(l)$ denotes the length of line segment $l$, and $\mu$ is the base length of the subparts. Note that when $k \leq 1$, we do not split the line segment. When dividing the line segment, the base length of subparts $\mu$ is determined by $\mu =$ input size $\times \epsilon$. We conduct an experiment to investigate the impact of ratio $\epsilon$ in Table D. Small ratio $\epsilon$ will split line segments into a shorter length while producing a greater number of subparts, and vice versa when using a large ratio $\epsilon$. As shown in Table D, although a small ratio $\epsilon$ produces a large number of augmented line segments, performance improvement is small. This is because the center and end points of small subparts are too close to each other to be distinguished, and thus become distractions for the model. Using a large ratio $\epsilon$ also shows small performance improvement because not only does the amount of augmented line segments decrease, but also these subparts result to resemble the original line segment. We observe the proper ratio $\epsilon$ is 0.125, which produces enough number of augmented line segments with different lengths and location from the originals.

When applying SoL augmentation, we split line segments into multiple subparts with overlapping portions with each other. To see the impact of retaining such overlap in SoL augmentation, we conduct an experiment as shown in Table E. W/o overlap shows a smaller performance boost than that of w/ overlap. Hence we conclude that using a larger number of augmented lines and preserving connectivity among subparts with overlaps can yield higher performance than without overlaps.

| M | Schemes | $F^H$ | $sAP^{10}$ | $LAP$ |
|---|---|---|---|---|
| 1 | Baseline | 74.3 | 48.9 | 48.1 |
| 2 | + Matching loss | 75.4 (+1.1) | 52.2 (+3.3) | 52.5 (+4.4) |
| 3 | + Line segmentation | 75.4 (0.0) | 52.9 (+0.7) | 53.7 (+1.2) |
| 4 | + Junction segmentation | 76.2 (+0.8) | 53.7 (+0.8) | 54.6 (+0.9) |
| 5 | + Length regression | 76.1 (-0.1) | 54.5 (+0.8) | 54.8 (+0.2) |
| 6 | + Degree regression | 76.2 (+0.1) | 55.1 (+0.6) | 55.3 (+0.5) |
| 7 | + SoL augmentation | **77.2** (+1.0) | **58.0** (+2.9) | **57.9** (+2.6) |

(a) Performance as training schemes accumulation. M denotes model number.

| Schemes | $F^H$ | $sAP^{10}$ | $LAP$ |
|---|---|---|---|
| Baseline (B) | 74.3 | 48.9 | 48.1 |
| B + Matching loss (M) | 75.4 (+1.1) | 52.2 (+3.3) | 52.5 (+4.4) |
| B + Geometric loss (G) | 75.0 (+0.7) | 50.7 (+1.8) | 51.3 (+3.2) |
| B + SoL augmentation (S) | 75.2 (+0.9) | 51.5 (+2.6) | 51.8 (+3.7) |
| B + M + G + S | **77.2** (+2.9) | **58.0** (+9.1) | **57.9** (+9.8) |

(b) Performance as training schemes added separately.

Table G: Extended ablation study of M-LSD-tiny on Wireframe. The baseline is trained with M-LSD-tiny backbone including only TP representation.

### C.4 Threshold of Matching Loss

In the matching loss, the threshold $\gamma$ decides whether to match the predicted and GT line segments. When $\gamma$ is small, the matching condition becomes strict, where the predicted line would be matched only with a highly similar GT line. When $\gamma$ is large, the matching condition becomes lenient, where the predicted line would be easily matched with the GT line even if it is not similar. We conduct an experiment to see the impact of the threshold $\gamma$ in matching loss. As shown in Table F, when the threshold is high ($\gamma \geq 10.0$), the matching condition is too broad, and poses a higher chance of predicted lines matching with non-similar GT lines. This becomes a distraction and shows performance degradation. On the other hand, when the threshold is too low ($\gamma = 2.5$), the matching condition is strict and consequently restrains the effect of the matching loss to be minor due to the small number of matched lines. We observe that a value around 5.0 is the proper threshold $\gamma$, which provides the optimal balance.

### C.5 Ablation Study of Geometric Loss

We conduct extended ablation experiments to analyze how each geometric information contributes to the model performance in Table G. Moreover, we include saliency map visualizations, which are generated from each feature map of geometric information as illustrated in Figure C.

**Line and Junction Segmentation.** In Table Ga, adding line and junction segmentation gives performance boosts in the following metrics: 0.8 in $F^H$, 1.5 in $sAP^{10}$ and 2.1 in $LAP$. Moreover, the junction and line attention on saliency maps of Figure Ca and Cb are precise, which shows that junction and line segmentations work as spatial attention cues for LSD.

**Length and Degree Regression.** In Table Ga, the line

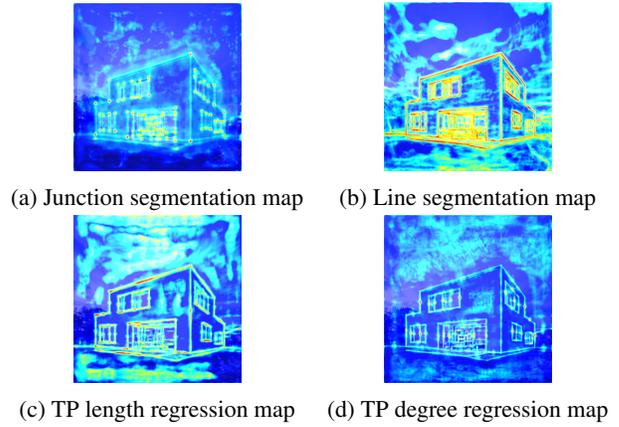

(a) Junction segmentation map  (b) Line segmentation map

(c) TP length regression map  (d) TP degree regression map

Figure C: Saliency maps generated from each feature map. M-LSD-tiny (M7 in Table Ga) model is used for generation.

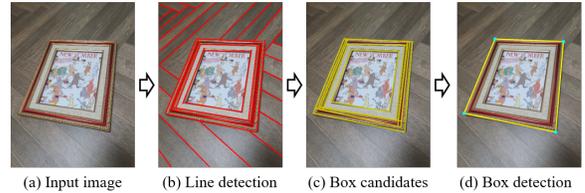

(a) Input image  (b) Line detection  (c) Box candidates  (d) Box detection

Figure D: Real-time box detection using M-LSD-tiny on a mobile device. Given an image as input to the mobile device as (a), line segments are detected using M-LSD-tiny as (b). Then, box candidates are computed from post-processing as (c), and finally we obtain box detection by a ranking process as (d).

prediction quality improves 1.4 in $sAP^{10}$ and 0.7 in $LAP$ by adding length and degree regression, while the pixel localization accuracy $F^H$ remains the same. The length saliency map in Figure Cc contains highlights on the entire line, and the degree saliency map in Figure Cd has highlights on the center points. We speculate that computing length needs the entire line information whereas computing the degree only needs parts of the line. Overall, learning with additional geometric information of line segments, such as length and degree, further increases the performance.

**Performance of Each Training Scheme** We conduct an additional ablation study by adding each training scheme to the baseline separately in Table Gb. The proposed training schemes in order of highest performance boost is matching loss, SoL augmentation and geometric loss. Overall, every training scheme gives a significant performance boost to the baseline.

### C.6 HAWP Line Segment Representation

We conduct an experiment using HAWP (Xue et al. 2020) line segment representation with our M-LSD backbones and training schemes. As shown in Table H, both M-LSD backbones and training schemes work well with HAWP line segment representation and produce competitive performance. However, the model parameters are relatively larger and the

| Setup | | | Wireframe | | | York | | | Params (M) | FPS |
|---|---|---|---|---|---|---|---|---|---|---|
| Backbone | Rep. | MLSD-s | $F^H$ | $sAP^{10}$ | $LAP$ | $F^H$ | $sAP^{10}$ | $LAP$ | | |
| M-LSD-tiny | HAWP | | 69.9 | 61.5 | 58.8 | 57.8 | 26.0 | 28.6 | **4.0** | **47.3** |
| M-LSD-tiny | HAWP | ✓ | 75.1 | 63.0 | 60.2 | 58.8 | 27.1 | 28.4 | **4.0** | **47.3** |
| M-LSD | HAWP | | 73.0 | 64.0 | 60.5 | 60.3 | 28.3 | 30.2 | 5.0 | 38.4 |
| M-LSD | HAWP | ✓ | **77.5** | **65.7** | **61.1** | **60.8** | **28.4** | **30.5** | 5.0 | 38.4 |

Table H: M-LSD with HAWP line segment representation (Rep.).

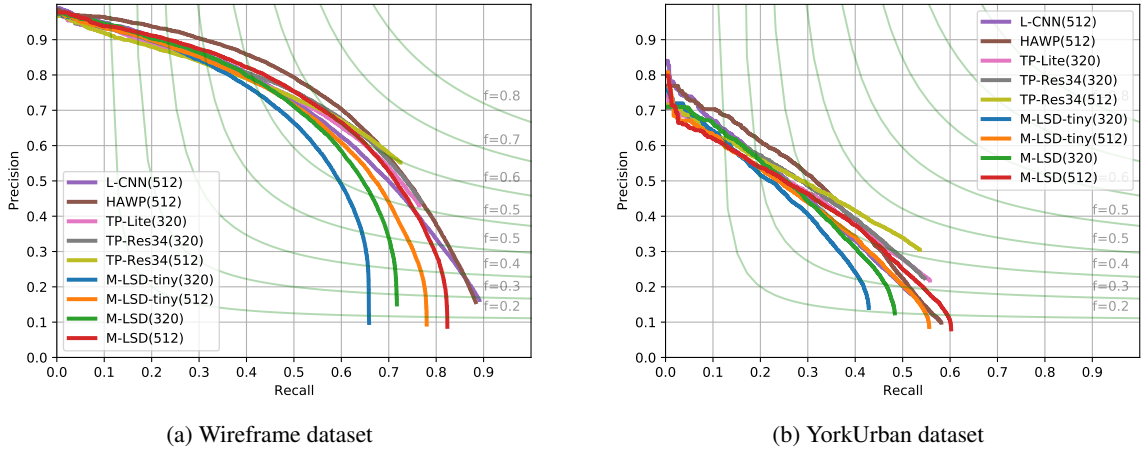

(a) Wireframe dataset    (b) YorkUrban dataset

Figure E: Precision-Recall (PR) curves of $sAP^{10}$ on Wireframe and YorkUrban datasets. (320) and (512) denote input image size.

FPS is low due to the complexity of the HAWP line segment representation.

### C.7 Applications

As line segments are fundamental low-level visual features, there are various real-world applications that use LSD. We show an example with real-time box detection on a mobile device as described in Figure D. We implement a box detector on a mobile device by using the M-LSD-tiny model. Since the application consists of line detection and post-processing, a model for the line detection has to be light and fast enough for real-time usage, when M-LSD-tiny is playing a sufficient role. The potential of real-time LSD on a mobile device can further be extended to other real-world applications like a book scanner, wireframe to image translation, and SLAM.

### C.8 Precision and Recall Curve

We include Precision-Recall (PR) curves of $sAP^{10}$ for L-CNN (Zhou, Qi, and Ma 2019), HAWP (Xue et al. 2020), TP-LSD (Huang et al. 2020), and M-LSD (ours). Figure E shows comparisons of PR curves on Wireframe and YorkUrban datasets.

### C.9 Visualization

We include more visualization results on Wireframe and YorkUrban datasets in Figure F. We compare our M-LSD model with AFM (Xue et al. 2019), L-CNN (Zhou, Qi, and Ma 2019), HAWP (Xue et al. 2020), TP-LSD-Res34 (Huang et al. 2020), LETR (Xu et al. 2021), and ground-truth. We use an input size of 512 for every method except that 320 is used for AFM, and the image is resized with the shortest side at least 1100 pixels for LETR.

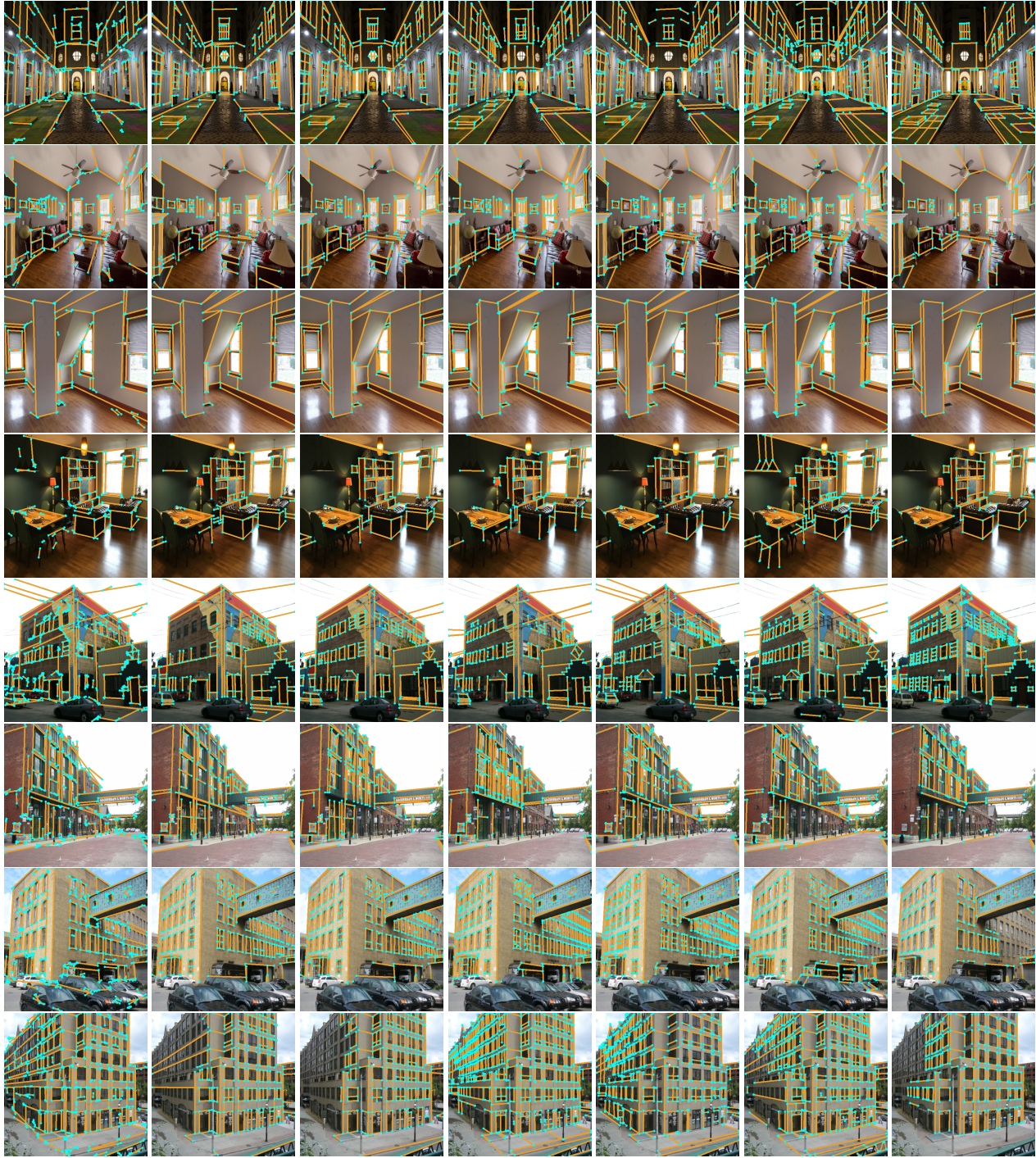

Figure F: Visualization of line segment detection methods. The columns are the results from AFM, LCNN, HAWP, TP-LSD-Res34, LETR, M-LSD (ours), and ground-truth. The top four rows are the results from Wireframe test set and bottom four rows are the results from YorkUrban test set.

# References

Almazan, E. J.; Tal, R.; Qian, Y.; and Elder, J. H. 2017. Mcmlsd: A dynamic programming approach to line segment detection. In *Proceedings of the IEEE Conference on Computer Vision and Pattern Recognition*, 2031–2039.

Burns, J. B.; Hanson, A. R.; and Riseman, E. M. 1986. Extracting straight lines. *IEEE transactions on pattern analysis and machine intelligence* (4): 425–455.

Cho, N.-G.; Yuille, A.; and Lee, S.-W. 2017. A novel linelet-based representation for line segment detection. *IEEE transactions on pattern analysis and machine intelligence* 40(5): 1195–1208.

Elder, J. H.; Almazàn, E. J.; Qian, Y.; and Tal, R. 2017. MCMLSD: A Probabilistic Algorithm and Evaluation Framework for Line Segment Detection. *Proceedings of the IEEE/CVF Conference on Computer Vision and Pattern Recognition* .

Furukawa, Y.; and Shinagawa, Y. 2003. Accurate and robust line segment extraction by analyzing distribution around peaks in Hough space. *Computer Vision and Image Understanding* 92(1): 1–25.

Huang, K.; Wang, Y.; Zhou, Z.; Ding, T.; Gao, S.; and Ma, Y. 2018. Learning to parse wireframes in images of man-made environments. In *Proceedings of the IEEE Conference on Computer Vision and Pattern Recognition*, 626–635.

Huang, S.; Qin, F.; Xiong, P.; Ding, N.; He, Y.; and Liu, X. 2020. TP-LSD: Tri-Points Based Line Segment Detector. *arXiv preprint arXiv:2009.05505* .

Kamat-Sadekar, V.; and Ganesan, S. 1998. Complete description of multiple line segments using the Hough transform. *Image and Vision Computing* 16(9-10): 597–613.

Matas, J.; Galambos, C.; and Kittler, J. 2000. Robust detection of lines using the progressive probabilistic hough transform. *Computer vision and image understanding* 78(1): 119–137.

Meng, Q.; Zhang, J.; Hu, Q.; He, X.; and Yu, J. 2020. LGNN: A Context-aware Line Segment Detector. In *Proceedings of the 28th ACM International Conference on Multimedia*, 4364–4372.

Sandler, M.; Howard, A.; Zhu, M.; Zhmoginov, A.; and Chen, L.-C. 2018. Mobilenetv2: Inverted residuals and linear bottlenecks. In *Proceedings of the IEEE conference on computer vision and pattern recognition*, 4510–4520.

Von Gioi, R. G.; Jakubowicz, J.; Morel, J.-M.; and Randall, G. 2008. LSD: A fast line segment detector with a false detection control. *IEEE transactions on pattern analysis and machine intelligence* 32(4): 722–732.

Xu, Y.; Xu, W.; Cheung, D.; and Tu, Z. 2021. Line segment detection using transformers without edges. In *Proceedings of the IEEE/CVF Conference on Computer Vision and Pattern Recognition*, 4257–4266.

Xu, Z.; Shin, B.-S.; and Klette, R. 2014. Accurate and robust line segment extraction using minimum entropy with Hough transform. *IEEE Transactions on Image Processing* 24(3): 813–822.

Xue, N.; Bai, S.; Wang, F.; Xia, G.-S.; Wu, T.; and Zhang, L. 2019. Learning attraction field representation for robust line segment detection. In *Proceedings of the IEEE/CVF Conference on Computer Vision and Pattern Recognition*, 1595–1603.

Xue, N.; Wu, T.; Bai, S.; Wang, F.; Xia, G.-S.; Zhang, L.; and Torr, P. H. 2020. Holistically-attracted wireframe parsing. In *Proceedings of the IEEE/CVF Conference on Computer Vision and Pattern Recognition*, 2788–2797.

Zhou, Y.; Qi, H.; and Ma, Y. 2019. End-to-end wireframe parsing. In *Proceedings of the IEEE/CVF International Conference on Computer Vision*, 962–971.



\end{document}